\documentclass[lettersize,journal]{IEEEtran}
\usepackage{amsmath,amsfonts}
\usepackage{algorithmic}
\usepackage{algorithm}
\usepackage{array}
\usepackage{textcomp}
\usepackage{stfloats}
\usepackage{subfigure}
\usepackage{url}
\usepackage{verbatim}
\usepackage{graphicx}
\usepackage{cite}
\usepackage{caption,subcaption}
\usepackage{hyperref}
\usepackage{bm}

\DeclareMathOperator*{\st}{s.t.}
\begin{document}

\title{Stackelberg Game Based Performance Optimization in Digital Twin Assisted Federated Learning over NOMA Networks}

\author{Bibo~Wu,~\IEEEmembership{Graduate Student Member, IEEE,}
	Fang~Fang,~\IEEEmembership{Senior Member, IEEE,}
	and Xianbin Wang,~\IEEEmembership{Fellow, IEEE}
	
	\thanks{An earlier version of this paper \cite{DTFL_VTC100} was presented in part at the 2024 IEEE 100th Vehicular Technology Conference (VTC2024-Fall), Washington DC, USA, 7-10 October 2024.
		
	Bibo Wu, Fang Fang and Xianbin Wang are with the Department of Electrical and Computer Engineering, and Fang Fang is also with the Department of Computer Science, Western University, London, ON N6A 3K7, Canada (e-mail: \{bwu293, fang.fang, xianbin.wang\}@uwo.ca).}
}

\maketitle

\begin{abstract}
Despite the advantage of preserving data privacy, federated learning (FL) still suffers from the straggler issue due to the limited computing resources of distributed clients and the unreliable wireless communication environment.
By effectively imitating the distributed resources, digital twin (DT) shows great potential in alleviating this issue.
In this paper, we leverage DT in the FL framework over non-orthogonal multiple access (NOMA) network to assist FL training process, considering malicious attacks on model updates from clients.
A reputation-based client selection scheme is proposed, which accounts for client heterogeneity in multiple aspects and effectively mitigates the risks of poisoning attacks in FL systems.
To minimize the total latency and energy consumption in the proposed system, we then formulate a Stackelberg game by considering clients and the server as the leader and the follower, respectively.
Specifically, the leader aims to minimize the energy consumption while the objective of the follower is to minimize the total latency during FL training.
The Stackelberg equilibrium is achieved to obtain the optimal solutions.
We first derive the strategies for the follower-level problem and include them in the leader-level problem which is then solved via problem decomposition.
Simulation results verify the superior performance of the proposed scheme.
\end{abstract}


\section{Introduction}
\IEEEPARstart{T}{he} rapid development of the Internet of Things (IoT) has fueled the explosive growth of mobile applications, leading to humongous amount of data available at edge devices \cite{YANG202233}.
In conventional edge learning methods, edge devices are required to upload raw data to a central server for processing.
This centralized mechanism not only consumes a substantial amount of communication resources but also increases the risk of leakage of private data.
Federated learning (FL), as a decentralized learning paradigm, is promising to address these issues.
The FL framework allows the server and all edge devices, also known as clients, to share a neural network.
After training the local model utilizing its own data, each client transmits the model parameters, e.g., gradients or weights, to the server for model aggregation \cite{FL_Mag}.
Therefore, the communication efficiency and data privacy can be significantly enhanced.
However, the performance of FL with synchronous aggregation is heavily restrained by the slow clients, i.e., the stragglers, which could arise from either insufficient local computing capability or undesirable communication environment.

To tackle the straggler issue by alleviating the computational burden on clients, mobile edge computing (MEC) is considered in FL systems \cite{MEC_FL, DDPGComplex}, where stragglers can offload computing tasks to the edge server, thereby reducing the required processing time for local model training.
However, this approach introduces extra offloading latency and still suffers from the undesirable communication environment.
In this context, digital twin (DT) technology offers an effective solution by eliminating the need for direct communication between clients and the edge server.
Specifically, DT enables real-time mapping of physical entities to their digital counterparts through software-defined frameworks, effectively capturing dynamic state information of physical entities and enhancing decision-making processes within networks \cite{DT_mag}.
Therefore, DT shows great potential in assisting FL model training by directly analyzing the DT network to get the physical entities' status information, hence simultaneously alleviating the impact of resource shortage at clients and unreliable communication on FL training performance.

Recent years have witnessed a surge in research focused on integrating DT with FL \cite{DTMEC_1, DTFL_JSAC1, IIoT, marine, LEC_2} and hierarchical federated learning (HFL) \cite{MEC_JSAC, DTFL_TVT} systems.
Specifically, aiming to maximizing the utility of FL services, the authors in \cite{DTMEC_1} adopted DT models to optimize the client selection and resource allocation problems by considering the dynamic bandwidth and clients.
In \cite{DTFL_JSAC1}, the authors proposed two communication-assisted sensing frameworks to enhance communication efficiency in DT-enabled mobile networks, including centralized and decentralized architectures of federated transfer learning (FTL).
The DT-empowered FL framework was applied to the industrial Internet of Things (IIoT) in \cite{IIoT}, aiming to enhance communication efficiency and reduce transmission energy cost.
By spreading model information via chaotic sequences, \cite{marine} minimized the total energy consumption while guaranteeing secrecy in FL-assisted marine DT Networks.
Non-orthogonal multiple access (NOMA) transmission was considered in \cite{LEC_2} to improve communication efficiency in the FL-enabled DT system, with a focus on minimizing energy consumption.
Based on the cloud-edge-client three layer structure, HFL is more complex compared to the conventional two-layer FL, thereby introducing more challenges when integrating it with DT.
In \cite{MEC_JSAC}, a DT-assisted resource scheduler was developed to optimize joint client scheduling and resource allocation in the HFL system, aiming to minimize the total communication and computation cost.
The authors in \cite{DTFL_TVT} integrated DT into the HFL framework over the heterogeneous cellular network, where the communication cost and failure rate were reduced through the DT network.

Client selection is another effective approach to alleviate the straggler issue in FL systems, which has been extensively studied in previous FL-related works, ranging from single-criterion-based \cite{FLCS1, FLCS2, cho2020client, FLage, FL_BBW} to multi-criteria-based schemes \cite{MR_CS2, fuzzy_ICC, HFL_NOMA_BBW}.
Specifically, by setting the updating time threshold for client selection, \cite{FLCS1} and \cite{FLCS2} investigated maximization problems of learning accuracy and convergence rate, respectively.
The authors in \cite{cho2020client} proposed a power-of-choice selection strategy, where clients with higher local training loss have high priority for selection at each FL round.
The metric of age of update, which accounts for the staleness of local models, was defined in \cite{FLage} and \cite{FL_BBW} to guide client selection in FL systems, and the latency minimization problem was studied by optimizing resource allocation.
Multi-criteria-based client selection schemes show significant advantages by accounting for multiple aspects of client heterogeneity. 
This approach guarantees fairness in client selection and is more practical for real-world applications.
In \cite{MR_CS2}, the heterogeneity of clients in computing capabilities, communication conditions, and available data sets was simultaneously considered for client selection, aiming to minimize the total energy cost for the long-term FL process.
A fuzzy logic-assisted client selection for HFL systems was proposed in \cite{fuzzy_ICC}, where clients' battery capacity, the distance between clients and the associated edge server,  and computational resources were jointly considered.
The authors in \cite{HFL_NOMA_BBW} also devised a fuzzy logic module for client selection, which integrates data quantity, channel quality, and model staleness, thereby effectively balancing client heterogeneity.
However, the aforementioned client selection schemes primarily focus on conventional factors and remain inadequate in ensuring FL performance, especially when confronted with the threat of malicious client attacks.

\subsection{Motivation and Contribution}
The training latency and energy consumption are major bottlenecks in DT-enabled FL systems.
Nevertheless, the aforementioned works have focused only on either latency or energy consumption minimization, or the cooperative interaction between them by introducing weight factors, while ignoring the competitive interaction between them so that fails to tradeoff the individual objectives of the server and clients.
Besides, the complexity of multi-criteria-based client selection designs is determined by the number of aspects considered for client heterogeneity. 
Existing works primarily focus on common aspects to accelerate the FL process, such as clients' computing capabilities and communication conditions, while neglecting the critical aspect of malicious attacks from clients, which can severely degrade FL performance if not carefully considered.
However, the deployment of DT in FL systems introduces new insights for client selection.
With real-time DT mapping, the status of clients becomes unnecessary for selection.
This allows for the incorporation of malicious attack considerations into client selection without increasing design complexity so that cater for unreliable training environment.

Motivated by the above analysis, this paper investigates the minimization problem of training latency and energy consumption in a DT-assisted FL system considering poisoning clients.
We harness the advanced computing ability and real-time mapping property of DT to assist FL training, thereby alleviating the straggler issue caused by insufficient resources of clients and dynamic wireless communication environment.
Besides, we leverage NOMA to improve spectrum efficiency in the system, which enables simultaneous transmission of local model parameters among multiple clients over the same channel.
To address the issue of unreliable clients, we first devise a reputation-based client selection scheme by evaluating the quality of model updates before global aggregation.
Thereafter, we construct a Stackelberg game to formulate the optimization problem, considering the competitive interaction between clients and the server, and the closed-form solutions are derived.
The main contributions of this paper are listed as follows.
\begin{enumerate}[]
	\item
	A DT-assisted FL system over NOMA network is studied to alleviate the straggler issue, where poisoning clients can launch malicious attacks on model updates, thereby degrading FL performance.
	A reputation-based, multi-criteria client selection scheme is proposed to address poisoning attacks by considering key factors such as clients' accuracy contributions, local model staleness, and the quality of interactions between clients and the server.
	
	\item
	A Stackelberg game is formulated to model the competitive interaction between clients and the server, where clients act as the leader and the server as the follower.
	Specifically, clients tend to minimize the total energy consumption via the optimization of mapping data ratio at DT and resource allocation, while the server aims to minimize the total latency to accelerate FL training with the optimal computation resource allocation provided by DT.
	
	\item
	To solve the formulated optimization problem, the Stackelberg equilibrium is analyzed.
	Firstly, two feasible strategies for the follower-level problem are derived based on the revealed insights.
	By incorporating these strategies into the leader-level problem, the problem decomposition approach is adopted to derive the closed-form solutions.
	Particularly, a Dinkelbach algorithm is proposed to address the optimization problem of transmitting power.
	
	
\end{enumerate}

\subsection{Organization}
The remaining content of this paper is organized as follows. 
The system model of DT-assisted FL over NOMA network is presented in Section II, followed by the reputation-based client selection design in Section III.
Section IV and Section V present the Stackelberg game-based problem formulation and derivation of optimal solutions, respectively.
Extensive simulations are included in Section VI and the conclusion of this paper is given in Section VII.

\section{System Model}

\begin{figure}[t]
	\centering
	\includegraphics[width=2.8in]{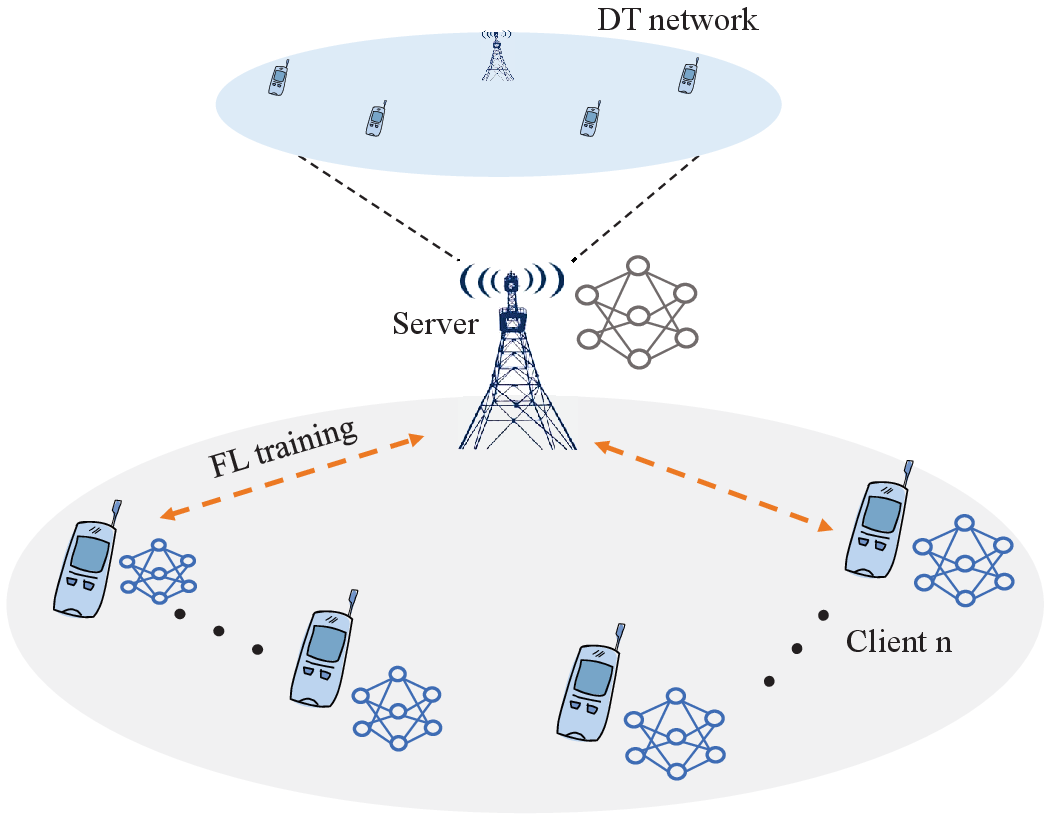}
	\caption{DT-assisted FL system model.}
	\label{SystemModel}
\end{figure}

Fig. \ref{SystemModel} shows the DT-assisted FL system model, which consists of a server $S$ deploying a DT network and $M$ clients indexed by ${\cal M} \in \left\{1, 2, ... , M \right\}$.
Due to the scarcity of communication resources, only $N$ clients ${\cal N} \in \left\{1, 2, ... , N \right\}$ are selected by the server to cooperatively train a global FL model at each round, and we assume $ {N \ll M}$.
Specifically, after receiving the global FL model broadcast from the server, each client performs local model training by utilizing its own data set ${{\cal{D}} _n} = \left\{ {\left( {{{\bf{x}}_i},{y_i}} \right)} \right\}_{i = 1}^{ {{D_n}} }$, where ${D_n}$ represents the number of data samples, ${{\bf{x}}_i}$ and ${y_i}$ denote the $i$-th data sample and its label, respectively.
Subsequently, the trained local model parameters are transmitted from selected clients to the server for model aggregation so that the global FL model is updated.
This process proceeds iteratively until the global FL model converges.

The DT network is deployed at the server $S$ to assist the FL training in the proposed system, which  builds real-time mapping between physical entities and digital objects.
In this sense, for the clients with inferior computation and communication resources, the model training process can be conducted under the assistance of DT network, hence reducing the latency of model training and transmission.
For client $n$, its DT model is denoted by ${DT}_n = \left\{ {\mathbf{w}_n}, {\hat {\cal D}_n} \right\}$, where ${\mathbf{w}_n}$ and ${\hat {\cal D}_n}$ are the local model parameter and the estimated data set of client $n$ with dimension $\hat {D}_n$, respectively. 
Different from previous DT assisted FL frameworks which assume the DT network can map the entire dataset of clients \cite{DTFL_TVT}, we consider that the DT network can only reflect the insensitive data of client to guarantee data privacy in our proposed framework.
Therefore, we set ${\hat {D}_n} \le {v_n^{\text{max}}}{D}_n + {\varepsilon } $, where ${v_n^{\text{max}}}$ denotes the maximum portion of the insensitive data at client $n$ and $\varepsilon $ indicates the deviation between the real data size and the estimated one via DT mapping.

\subsection{FL Training}
Define ${{l}\left( {{\bf{w}}_n^t,{{\bf{x}}_i},{y_i}} \right)}$ as the local loss function of client $n$, where ${\bf{w}}_n^t$ denotes the local model parameter at the $t$-iteration of FL.
With the partial local model training and the assistance of DT network deployed at the server, the entire local model can be trained at the server. 
Defined $v_nD_n$ as the size of data represented at the DT network, where $v_n \in [0, {v_n^{\text{max}}}]$ indicates the portion of mapping data.
Therefore, the data size at client $n$ for local model training is $\left( {1 - {v_n}} \right){D_n}$.
The total loss of client $n$ can be written as
\begin{equation}\label{}
	\begin{aligned}
		{L}\left( {\bf{w}}_n^t \right) = \frac{1} {{\left( {1 - {v_n}} \right){D_n}}} \sum\limits_{i = 1}^{{D_n}} {{l}\left( {{\bf{w}}_n^t,{{\bf{x}}_i},{y_i}} \right)}.
	\end{aligned}
\end{equation}
Client $n$ updates its local model at the $t$-th iteration via the following stochastic gradient descent (SGD) manner:
\begin{equation}\label{GD}
	\begin{aligned}
		{\bf{w}}_n^t = {\bf{w}}_n^{t-1} - \eta \nabla {L}\left( {{\bf{w}}_n^{t - 1}} \right),
	\end{aligned}
\end{equation}
wherein $\eta$ denotes the learning rate.
Subsequently, the server aggregates selected clients' local model parameters and the model parameters ${\mathbf{w}}_S^t$ generated from the DT network, and then the global model can be expressed as 
\begin{equation}\label{GM}
	\begin{aligned}
		{\bf{w}}^t = \frac{1}{D} \left( {\sum\limits_{n \in {\cal N}} {\left[ {\left( {1 - {v_n}} \right){D_n}{\mathbf{w}}_n^t + ({v_n}{D_n} + {\varepsilon }){\mathbf{w}}_S^t} \right]} } \right),
	\end{aligned}
\end{equation}
where $D = \sum \nolimits_{n = 1}^{N} D_n$ denotes the total data size of all selected clients.
After that, the aggregated global model is broadcast by the server to all clients and the above process repeats until FL convergence.

The convergence of the global FL model ${\bf{w}}^t$ can be proved as follows.
In the proposed DT-assisted FL system, the server $S$ can assist FL training by utilizing the deployed DT network.
It can be regarded as a powerful client that engages in the original FL process, and thus the convergence analysis is similar with the centralized gradient decent method \cite{DTFL_TVT}.
After global model aggregation and broadcast, we can obtain ${\mathbf{w}}_n^t = {\mathbf{w}}_S^t = {\mathbf{w}}^t$.
According to \eqref{GD}, we can rewrite \eqref{GM} as
 \begin{equation}\label{}
 	\begin{aligned}
 			{{\bf{w}}^t} & = \frac{1}{D}\left( {\sum\limits_{n \in {{\cal N}}} {\left[ {\left( {1 - {v_n}} \right){D_n}{\bf{w}}_n^t + ({v_n}{D_n} + \varepsilon ){\bf{w}}_S^t} \right]} } \right)\\
 			&= \frac{1}{D}\sum\limits_{n \in {{\cal N}}} {\left( {\left( {1 - {v_n}} \right){D_n}\left( {{{\bf{w}}^{t - 1}} - \eta \nabla L\left( {{{\bf{w}}^{t - 1}}} \right)} \right)} \right)} \\
 			& \quad + \frac{1}{D}\sum\limits_{n \in {{\cal N}}} {\left( {({v_n}{D_n} + \varepsilon )\left( {{{\bf{w}}^{t - 1}} - \eta \nabla L\left( {{{\bf{w}}^{t - 1}}} \right)} \right)} \right)} \\
 			&= \frac{1}{D}\sum\limits_{n \in {{\cal N}}} {\left( {{D_n}{{\bf{w}}^{t - 1}} - {D_n}\eta \nabla L\left( {{{\bf{w}}^{t - 1}}} \right)} \right)} \\
 			&\quad + \frac{1}{D}\sum\limits_{n \in {{\cal N}}} {\left( {\varepsilon {{\bf{w}}^{t - 1}} - \varepsilon \eta \nabla L\left( {{{\bf{w}}^{t - 1}}} \right)} \right)} \\
 			&= \Gamma \left( {{{\bf{w}}^{t - 1}} - \eta \nabla L\left( {{{\bf{w}}^{t - 1}}} \right)} \right),
 	\end{aligned}
 \end{equation}
where $\Gamma  = 1 + \frac{{\varepsilon {N}}}{D}$.
Thus, following the gradient descent property, the global model ${{\bf{w}}^t}$ can converge to the expected accuracy.

\subsection{Computing Model}
The computing model in the proposed system consists of local computing at clients as well as DT computing at the server.
Define $f_n$ as client $n$'s CPU frequency to conduct local model training.
The local computing latency of client $n$ is calculated as
\begin{equation}\label{LocalTime}
	\begin{aligned}
		t_n^{\text{cmp}} = \frac{{{c_n}\left( {1 - {v_n}} \right){D_n}}}{{{f_n}}},
	\end{aligned}
\end{equation}
where $c_n$ is the required CPU cycles for processing one data unit.
The corresponding energy consumption is 
\begin{equation}\label{LocalEnergy}
	\begin{aligned}
		e_n^{\text{cmp}} = \frac{\tau }{2}{c_n}\left( {1 - {v_n}} \right){D_n}f_n^2,
	\end{aligned}
\end{equation}
where ${\tau } = 2\times10^{-28}$ represents the effective capacitance coefficient \cite{LocComEn}.

Regarding the DT computing at the server, denote the server's frequency by $f_S$, and the computing latency can be expressed as 
\begin{equation}\label{}
	\begin{aligned}
		t_n^S = \frac{{{c_n}{{\hat D}_n}}}{{{\alpha _n}{f_S}}} = \frac{{{c_n}\left( {{v_n}{D_n} + \varepsilon } \right)}}{{{\alpha _n}{f_S}}},
	\end{aligned}
\end{equation}
where ${\alpha _n} \in [0, 1]$ indicates the frequency coefficient allocated to client $n$ for the computation of mapping data.
Due to the sufficient power of the server, we reasonably ignore the corresponding energy consumption, which is commonly adopted in previous works \cite{DT_MEC, DT_IoT}.

\subsection{Communicating Model}
We consider NOMA transmission to improve the communication efficiency between the server and selected clients.
By adopting the superposition coding at the transmitter, NOMA enables multiple clients to transmit their local model parameters through the same channel simultaneously.
Thus, one direct benefit is that more clients can participate in global model aggregation at each FL round given the limited communication resources.
However, this technique inevitably introduces interference among clients' signals, which can be effectively addressed by utilizing successive interference cancellation (SIC) at the receiver \cite{Ding_Survey_2017}.
For ease of analysis, we assume the receiver, i.e., the server $S$, is powerful enough to conduct SIC for multiple clients in this paper.
Denote the encapsulated signal transmitted from client $n$ by $s_n$ which is normalized as $\mathbb{E}(|s_n|^2) = 1$ \cite{FLRIS}.
Define $p_n$ and $h_n$ as client $n$'s transmitting power and channel gain between client $n$ and the server, respectively.
In this case, the superposition signal received at the server is expressed as
\begin{equation}\label{}
	\begin{aligned}
		{y} = \sum\limits_{n = 1}^{N} {\left( {\sqrt {{p_n}} {h_{n}}{s_n}} \right)}  + {{\hat n}},
	\end{aligned}
\end{equation}
where ${{\hat n}} \sim {\cal{CN}} \left( {0,{\sigma_n ^2}} \right)$ is the additive white Gaussian noise (AWGN) with a mean of $0$ and a variance of ${\sigma_n ^2}$.

In the uplink NOMA transmission, SIC is performed at the receiver according to the following criteria.
Firstly, the receiver decodes the strongest signal by treating other signals as interference.
Subsequently, the decoded signal is subtracted from the superposition signal and the second strongest signal is then decoded.
This process repeats until the last signal is decoded successfully without any interference.
Note that the design of decoding order is complicated and beyond the scope of this research.
Thus, in this paper, we simply assume the decoding order at the server aligns with the descending order of channel gains, i.e., ${\left| {{h_{1}}} \right|^2}  \ge {\left| {{h_{2}}} \right|^2} \ge  \cdots  \ge {\left| {{h_{{N}}}} \right|^2}$.
Thus, the achievable transmission rate of client $n$ can be calculated by 
\begin{equation}\label{}
	\begin{aligned}
		{R_n} = B{\log _2}\left( {1 + \frac{{{p_n}{{\left| {{h_n}} \right|}^2}}}{{\sum\limits_{j = n + 1}^N {{p_j}{{\left| {{h_j}} \right|}^2}}  + \sigma _n^2}}} \right),
	\end{aligned}
\end{equation}
where $B$ is the available channel bandwidth.
Particularly, for the last client $N$ which is decoded without interference, its achievable data rate is written as ${R_N} = B{\log _2}\left( {1 + \frac{{{p_N}{{\left| {{h_N}} \right|}^2}}}{{\sigma _N^2}}} \right)$.

Given the above data rate, the latency and energy consumption for transmitting local model parameter of client $n$ can be respectively expressed as
\begin{equation}\label{}
	\begin{aligned}
		t_n^{\text{com}} = \frac{{{d_n}}}{{{R_n}}},
	\end{aligned}
\end{equation} 

\begin{equation}\label{}
	\begin{aligned}
		e_n^{\text{com}} = {p_n}{t_n^{\text{com}}},
	\end{aligned}
\end{equation}
where $d_n$ is the size of local model parameters which is assumed to be unique among all clients due to local models have similar numbers of elements \cite{FLT1}.


\section{Reputation-based Client Selection Design}
In this section, the reputation-based client selection scheme is devised to achieve reliable FL considering poisoning clients.
Generally, high-reputation clients contribute more to the global model update, thereby accelerating FL convergence.
In this context, we identify the reputation of clients from three different aspects: accuracy contribution, model staleness and positive interactions.
Note that it can also be regarded as a multi-criteria-based client selection strategy, which effectively balances client heterogeneity in multiple dimensions, as investigated in \cite{MR_CS2, fuzzy_ICC, HFL_NOMA_BBW}.
However, these works did not consider the unreliable clients which may cause malicious attack on the global model update.
In contrast, we evaluate the quality of local model updates before global model aggregation to avoid selecting the unreliable clients that may cause a decline in FL performance.
The details of the reputation-based client selection scheme are presented as follows.

\subsubsection{Accuracy contribution (AC)}
The quantity of training data plays a significant role in improving FL performance.
There is a consensus that the larger size of training data contributes to the improvement of FL model accuracy.
However, this relationship is not linear \cite{Incentive_IoT}.
This indicates that solely considering data quantity of clients is insufficient for client selection, which, however, is a common design in previous studies.
To address this issue, we adopt AC as a metric to describe the relationship between the global model accuracy and the training data of clients. 
Mathematically, the AC value of client $n$ can be expressed by the following Weibull model \cite{Weibull}:
\begin{equation}\label{}
	\begin{aligned}
		A{C_n} = \varpi _n^1 - \varpi _n^2\exp \left[ { - \varpi _n^3\left( {{D_n} + \varepsilon } \right)} \right],
	\end{aligned}
\end{equation}
where $\varpi _n^i$ is the predefined parameter and ${D_n} + \varepsilon$ is derived from $ \left( {1 - {v_n}} \right){D_n} + {{\hat D}_n} = \left( {1 - {v_n}} \right){D_n} + {v_n}{D_n} + {\varepsilon }$ which accounts for the DT mapping deviation in training data.
Note that $A{C_n}$ is an increasing concave function, indicating that while the global model accuracy improves with an increase in the size of training data, the rate of improvement decreases.
Therefore, we cannot select clients based solely on their training data quantity, as a larger quantity of training data also introduces higher latency and energy consumption in FL process.
The adoption of AC metric can effectively balance the tradeoff between learning accuracy and the required cost.

\subsubsection{Model staleness (MS)} 
This factor measures the elapsed time since the client's most recent selection by the server, also termed the age of update in \cite{FL_BBW}. 
MS is crucial for client selection in FL systems, as stale local models negatively impact the global model convergence, as revealed in \cite{AoI}.
Thus, to achieve better FL performance, the average MS value across the entire system should be kept as low as possible.
In other words, client selection at the server should ensure that local models are updated in a timely manner.
Define the MS value of client $n$ at the $t$-th FL round as $M{S_n^t}$, which is calculated by
\begin{equation}\label{AoU}
	\begin{aligned}
		M{S_n^t} = \left\{ {\begin{array}{*{20}{l}}
				{M{S_n^{t-1}} + 1,{\text{ }}a_{n}^{t - 1} = 0}, \\ 
				{1,{\text{ }}a_{n}^{t - 1} = 1}, 
		\end{array}} \right.
	\end{aligned}
\end{equation}	
where $a_{n}^{t - 1}$ denotes the selection indicator of client $n$ at the $(t - 1)$-th FL round, i.e., if client $n$ is selected by the server, $a_{n}^{t - 1} = 1$; otherwise $a_{n}^{t - 1} = 0$.
This definition shows that the MS value of any client depends on its selection status in the previous round.
Note that a larger MS value of a client indicates a more informative model update, which promotes the convergence of the global FL model and increases the probability for client selection.
At any round $t$, the normalized MS value is defined as follows:
\begin{equation}\label{}
	\begin{aligned}
		\overline {MS}_n^t = \frac{{MS_n^t}}{{\sum\nolimits_{n'}^N {MS_{n'}^t} }}.
	\end{aligned}
\end{equation}	

\subsubsection{Positive interactions (PI)}
In general, the server is reluctant to select unreliable clients for global model aggregation, as they may launch malicious model updates that degrade FL performance.
Thus, the quality of local model updates needs to be evaluated before model aggregation using various poisoning attack detection schemes.
PI indicate that local model updates contribute to the increase in global model accuracy.
Conversely, negative interactions (NI) lead to a decline in global model accuracy.
Note that the design of attack detection algorithm is beyond the scope of this paper, and we simply adopt the classical reject on negative influence (RONI) scheme proposed in \cite{PI} to detect PI and NI during the FL process.
Specifically, by comparing the influence with and without a specific local model update on a predefined dataset, the quality of this local model update can be validated.
If it degrades the training performance beyond a specified threshold, this local model update will not be considered for global model aggregation and is recorded as NI by the server.
Otherwise, the local model update is regarded as PI.
For client $n$, its PI degree can be calculated by
\begin{equation}\label{}
	\begin{aligned}
		P{I_n} = \frac{I_n^{PI}} {I_n^{PI} + I_n^{NI}},
	\end{aligned}
\end{equation}	
where $I_n^{PI}$ and $I_n^{NI}$ are the recorded number of PI and NI between client $n$ and the server, respectively.

By jointly considering the above three factors, the reputation value of client $n$ can be expressed in a weighted manner as follows
\begin{equation}\label{}
	\begin{aligned}
		Z_n = {\xi _1}{AC_n} + {\xi _2}{\overline {MS}_n} + {\xi _3}{PI_n},
	\end{aligned}
\end{equation}	
where ${\xi _i}$ is the weight of each factor and $\overline {MS}_n$ is the simplified form of $\overline {MS}_n^t$.
We assume that the server has knowledge of all clients' reputation values prior to selecting clients.
In this case, at each FL round, all clients can be sorted in a descending order based on their reputation values, and the first $N$ clients are selected by the server to engage in global model aggregation.

\section{Stackelberg Game-based Problem Formulation}
Given the reputation-based client selection scheme, the system-wise delay and energy consumption under synchronous aggregation mechanism at the server can be respectively expressed as
\begin{equation}\label{}
	\begin{aligned}
		T = \max \left\{   {t_n^{\text{cmp}} + t_n^{\text{com}}, t_n^{S}}   \right\}, n \in {\cal N},
	\end{aligned}
\end{equation}
\begin{equation}\label{}
	\begin{aligned}
		E = \sum\limits_{n \in {\cal N}} {\left( {e_n^{\text{cmp}} + e_n^{\text{com} }}\right)}.
	\end{aligned}
\end{equation}
To balance the inherent trade-off between latency and energy consumption in the proposed DT-assisted FL system, we investigate the competitive interaction between them by treating the minimization of latency and energy consumption as the objectives of the server and clients, respectively.
Specifically, for the resource-constrained clients, they tend to minimize the total energy consumption to save resources via the optimization of mapping data ratio and resource allocation.
For the server, it aims to minimize the total latency to accelerate FL training without concerning the energy consumption owing to its sufficient power.
In addition, clients can give resource allocation scheme firstly by predicting the possible strategy, and then the server decides its own strategy based on the observed strategy of clients.
This competitive process aligns with the Stackelberg game by treating clients as the leader and the server as the follower \cite{Game_Kaidi}.
It can also be regarded as an incentive mechanism for clients and the server in FL training, as their individual objectives are considered and optimized.
Next, the Stackelberg game-based problem is formulated, where the leader-level problem and the follower-level problem are investigated.

\subsection{Leader-Level Problem}
The leader in the Stackelberg game, i.e., clients, aims to minimize the total energy consumption during FL training.
The leader-level problem can be formulated as 
\begin{subequations}\label{llp}
	\begin{align}
		\mathop {\min }\limits_{{\bf{p}},{\bf{f}}, {\bf{v}}} \quad & E \label{} \\
		\st\ \quad  & t_n^{\text{cmp}} + t_n^{\text{com}} \le  {T^{\max}}, \forall n \in {\cal N}, \label{llp1}\\
		& t_n^{S}  \le  {T^{\max}}, \forall n \in {\cal N}, \label{llp2}\\
		& p_n^{\min}  \le  {p_n} \le p_n^{\max },\forall n \in {\cal N}, \label{llp4}\\
		& f_n^{\min}  \le {f_n} \le f_n^{\max },\forall n \in {\cal N}, \label{llp5}\\
		& 0 \le v_n \le {v_n^{\text{max}}}, \forall n \in {\cal N}. \label{llp6}
	\end{align}
\end{subequations}
Constraints \eqref{llp1} and \eqref{llp2} are the maximum latency limit for FL training and DT computing, respectively.
Constraints \eqref{llp4} and \eqref{llp5} limit the range of transmitting power and local computing frequency for each client, respectively.
In \eqref{llp6}, the range of values for the portion of mapping data at the DT network is limited by the maximum portion of the insensitive data at each client.

\subsection{Follower-Level Problem}
In this Stackeberg game, the server is regarded as the follower, which aims to minimize the total training latency to accelerate the convergence of FL.
The follower-level problem is formulated as
\begin{subequations}\label{flp}
	\begin{align}
		\mathop {\min }\limits_{{\bm{\alpha }}} \quad & T \label{} \\
		\st\ \quad & 0 \le \alpha_n \le 1, \forall n \in {\cal N}, \label{flp1}\\
		& \sum\limits_{n \in {\cal N}} {{\alpha _n}}  \le 1, \label{flp2}
	\end{align}
\end{subequations}
where \eqref{flp1} and \eqref{flp2} constraint the value range and the total limit of the frequency coefficients allocated to selected clients at the server, respectively.

In the formulated Stackelberg game-based problem, the leader (all selected clients) first decides the mapping data ratio and the resource allocation strategy to minimize total energy consumption.
Subsequently, based on the observation towards the leader, the follower (the server) determines the allocated frequency coefficients to minimize the total training latency.
The leader can obtain its optimal strategy by predicting the follower's best response.
In this sense, the optimal solutions of \eqref{llp} and \eqref{flp} can be obtained by achieving the Stackelberg equilibrium as follows 
\begin{equation}\label{}
	\begin{aligned}
		& F_{\text {leader}}({\bf{p}}^*,{\bf{f}}^*, {\bf{v}}^*, {\bm{\alpha }}^*) \le F_{\text {leader}}({\bf{p}},{\bf{f}}, {\bf{v}}, {\bm{\alpha }}^*), \\
		& F_{\text {follower}}({\bf{p}}^*,{\bf{f}}^*, {\bf{v}}^*, {\bm{\alpha }}^*) \le F_{\text {follower}}({\bf{p}}^*,{\bf{f}}^*, {\bf{v}}^*, {\bm{\alpha }}),
	\end{aligned}
\end{equation}
where $F_{\text {leader}}$ and $F_{\text {follower}}$ denote the objective functions of the leader and the follower, respectively.
To derive the Stackelberg equilibrium, the optimal solution for the follower-level problem needs to be derived firstly, based on which the leader-level problem is solved optimally.

\section{Proposed Solutions}
In this section, we first address the follower-level problem by converting it to a more tractable form based on mathematical analysis.
Next, leveraging the optimal response from the follower-level problem, we decompose the leader-level problem into several subproblems, which are then solved iteratively to achieve optimal solutions for the original problem.
Finally, the overall algorithm is designed to achieve the Stackelberg equilibrium.

\subsection{Solution of Follower-Level Problem}

According to the principle of pure NOMA transmission, the signals of selected clients are transmitted to the server simultaneously as a superposition signal.
This indicates that all selected clients start the NOMA transmission procedure at the same time instant.
By ignoring the consumed time of SIC at the server side, the transmission time for all selected clients is assumed to be identical \cite{NOMA_time}, which can be expressed as 
\begin{equation}\label{t_com}
	\begin{aligned}
		t_n^{\text{com}} = t^{\text{com}}, \forall n \in {\cal N}.
	\end{aligned}
\end{equation}
In this case, it is required that the selected clients complete their local model training within a time period $t^{\text{cmp}}$ before the NOMA transmission phase.
By integrating \eqref{LocalTime} into \eqref{LocalEnergy}, the energy consumption of client $n$ for local training can be rewritten as
\begin{equation}\label{LocalEnergy1}
	\begin{aligned}
		e_n^{\text{cmp}} = \frac{\tau }{2}{c_n}\left( {1 - {v_n}} \right){D_n}f_n^2 = \frac{{\tau {{\left[ {{c_n}\left( {1 - {v_n}} \right){D_n}} \right]}^3}}}{{2{{\left( {t_n^{\text{cmp}}} \right)}^2}}}.
	\end{aligned}
\end{equation}
From \eqref{LocalEnergy1}, it is observed that there is an inverse proportionality between $e_n^{\text{cmp}}$ and ${t_n^{\text{cmp}}}$.
Note that the selected clients are forced to complete their local model training within $t^{\text{cmp}}$.
Thus, aiming to minimize the energy consumption for local training, all selected clients tend to maximize the local computing time to $t^{\text{cmp}}$, which leads to 
\begin{equation}\label{t_cmp}
	\begin{aligned}
		t_n^{\text{cmp}} = t^{\text{cmp}}, \forall n \in {\cal N}.
	\end{aligned}
\end{equation}

Given the above analysis, the follower-level problem \eqref{flp} can be transformed as 
\begin{subequations}\label{flp_1}
	\begin{align}
		\mathop {\min }\limits_{{\bm{\alpha }}} & \max \left\{   {t^{\text{total}}, t_n^{S}}   \right\} \label{} \\
		\st\ & \eqref{flp1}, \eqref{flp2},
	\end{align}
\end{subequations}
where $t^{\text{total}} = {t^{\text{cmp}} + t^{\text{com}}}$ is regarded as a constant.
To obtain the optimal solution of $\alpha_n$, we first propose the following Theorem 1.

~\\
{\noindent\bf{Theorem 1}} {\textit{According to the principle of avoiding resource waste and utilizing full computing resources at the server, the optimal solution $\alpha_n^*$ is derived when $t_1^S = t_2^S = \cdots = t_N^S = t^S$, where $t^S \ge t^{\text{total}}$.
} 

~\\
{\noindent\textit{Proof.}}
\begin{figure}[t]
	\centering
	\subfigure[Case 1: $\max \left\{   { t_n^{S}}  \right\} \le t^{\text{total}}$] { \label{Theorem1_a}
		\includegraphics[width=0.44\columnwidth]{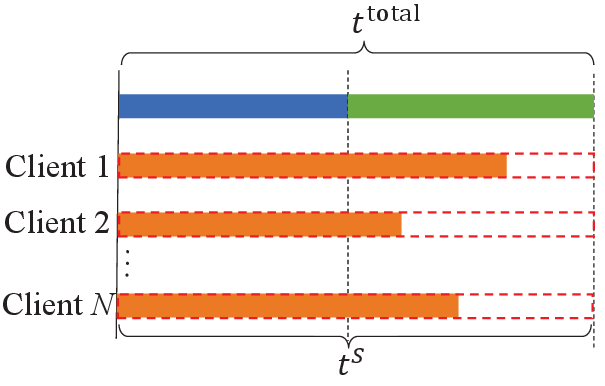}
	}
	\subfigure[Case 2: $\max \left\{   { t_n^{S}}  \right\}  >  t^{\text{total}}$] { \label{Theorem1_b}
		\includegraphics[width=0.48\columnwidth]{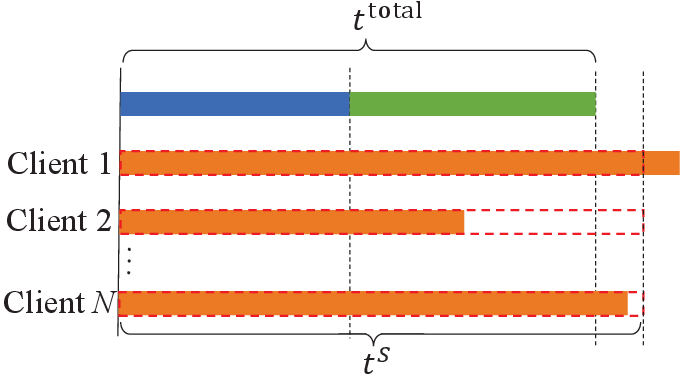}
	}
	\caption{Analysis of $t^{\text{total}}$ and $t_n^{S}$. }
	\label{TimeAnalysis}
\end{figure}
The problem \eqref{flp_1} can be interpreted as finding the optimal solution $\alpha_n^*$ which minimizes the maximum value between $t^{\text{total}}$ and $t_n^{S}$, subject to constraints on the value limit and total budget.
Given that $t^{\text{total}}$ is regarded as a constant, the optimization of problem \eqref{flp_1} is determined by $\max \left\{   { t_n^{S}} {|\forall n \in {\cal N}}  \right\}$, which is simplified as $\max \left\{   { t_n^{S}}  \right\}$ in the following analysis.
We present the proof based on two cases: $\max \left\{   { t_n^{S}}  \right\} \le t^{\text{total}}$ and $\max \left\{   { t_n^{S}}  \right\}  >  t^{\text{total}}$, as shown in Fig. \ref{TimeAnalysis}\subref{Theorem1_a} and Fig. \ref{TimeAnalysis}\subref{Theorem1_b} respectively.
Case 1 indicates that the server is able to provide sufficient computing resources for each selected client to process the DT mapping data.
However, it is not an efficient solution as the problem \eqref{flp_1} is determined by the constant time $t^{\text{total}}$ in this case.
Thus, to avoid wasting resources, the server tends to allocate the minimum total computing resources for clients, which leads to $t_1^S=t_2^S=\cdots=t_N^S=t^S=t^{\text{total}}$.
For the case 2, the server would utilize its full computing resources to decrease the value of $\max \left\{   { t_n^{S}}  \right\}$.
Additionally, the server tends to sacrifice the computing resources of clients with less DT computing time to compensate for the clients with larger DT computing time until reaching a balance of $t_1^S = t_2^S = \cdots = t_N^S = t^S$.
By utilizing the full computing resources of the server, i.e., $\sum\nolimits_{n \in {\cal N}} {{\alpha _n}}  = 1$, if $t^S < t^{\text{total}}$, the same analysis of case 1 can be utilized so that $t^S=t^{\text{total}}$.
Otherwise, it indicates that the server cannot provide sufficient computing resources for selected clients to process the DT mapping data, and hence $t^S > t^{\text{total}}$.
This completes the proof of Theorem 1.

Based on Theorem 1, we can derive the optimal solution $\alpha_n^*$ from following two cases.
\subsubsection{When $t^S=t^{\text{total}}$}
Note that this case exits if and only if $\sum\nolimits_{n \in {\cal N}} {{\alpha _n}}  \le 1$ is satisfied.
Otherwise, the server will decrease the computing resources allocated to each client until satisfy the total budget constraint.
Consequently, the DT computing time of each client increases, which transfers case 1 into case 2.
According to $t_n^S=t^S=t^{\text{total}}, \forall n \in {\cal N}$, we can directly derive the optimal solution $\alpha_n^*$ as
\begin{equation}\label{Solution_a}
	\begin{aligned}
		\alpha_n^* = \frac{{{c_n}{{\hat D}_n}}}{{{t^{\text{total}}}{f_S}}}.
	\end{aligned}
\end{equation}

\subsubsection{When $t^S>t^{\text{total}}$}
In this case, $\sum\nolimits_{n \in {\cal N}} {{\alpha _n}}  = 1$ is always satisfied, which indicates the full computing resources are utilized at the server.
Given $t_n^S=t^S, \forall n \in {\cal N}$, we have 
\begin{equation}\label{}
	\begin{aligned}
		\frac{{{c_1}{{\hat D}_1}}}{{{\alpha _1}{f_S}}} = \frac{{{c_2}{{\hat D}_2}}}{{{\alpha _2}{f_S}}} =  \cdots  = \frac{{{c_N}{{\hat D}_N}}}{{{\alpha _N}{f_S}}} = t^{S}.
	\end{aligned}
\end{equation}
By defining ${e_n} = \frac{{{a_n}}}{{{a_1}}} = \frac{{{b_n}}}{{{b_1}}}$, we can equivalently write $\frac{{{a_1}}}{{{b_1}}} = \frac{{{a_2}}}{{{b_2}}} =  \cdots  = \frac{{{a_N}}}{{{b_N}}}$ as $\frac{{{a_1}}}{{{b_1}}} = \frac{{{e_2}{a_1}}}{{{e_2}{b_1}}} =  \cdots  = \frac{{{e_N}{a_1}}}{{{e_N}{b_1}}}$.
Thereafter, $\frac{{{a_1}}}{{{b_1}}} = \frac{{{a_1}\left( {1 + {e_2} + {e_3} +  \cdots  + {e_N}} \right)}}{{{b_1}\left( {1 + {e_2} + {e_3} +  \cdots  + {e_N}} \right)}} = \frac{{{a_1} + {a_2} +  \cdots  + {a_N}}}{{{b_1} + {b_2} +  \cdots  + {b_N}}}$.
Thus, for client $n$, we have
\begin{equation}\label{}
	\begin{aligned}
		\frac{{{c_n}{{\hat D}_n}}}{{{\alpha _n}{f_S}}} = \frac{{\sum\nolimits_n^N {{c_n}{{\hat D}_n}} }}{{\sum\nolimits_n^N {{\alpha _n}{f_S}} }}.
	\end{aligned}
\end{equation}
Due to $\sum\nolimits_{n \in {\cal N}} {{\alpha _n}}  = 1$, we can derive the optimal solution as
\begin{equation}\label{OptAl}
	\begin{aligned}
		\alpha_n^* = \frac{{{c_n}{{\hat D}_n}}}{{\sum\nolimits_n^N {{c_n}{{\hat D}_n}} }}.
	\end{aligned}
\end{equation}

\subsection{Solution of Leader-Level Problem}
It is revealed in \cite{EE_time} that as the NOMA transmission time increases, the energy consumption of users decreases monotonously.
Therefore, before solving the leader-level problem \eqref{llp}, we first present the following Remark 1.

~\\
{\noindent\bf{Remark 1}} {\textit{Given any strategy of the follower, the leader tends to increase the transmission time to minimize the energy consumption by reducing the transmitting power.
} 
~\\

Note that $ t^{\text{total}} \le t^S$ is always satisfied as stated in Theorem 1, where $t^{\text{total}} = {t^{\text{cmp}} + t^{\text{com}}}$.
According to Remark 1, to minimize energy consumption, the leader, i.e., the selected clients, tends to increase the transmission time $t^{\text{com}}$ until $ t^{\text{total}} = t^S$ is hold.
Given this conclusion, the constraint \eqref{llp2} in the leader-level problem \eqref{llp} can be removed, and we can rewrite the problem \eqref{llp} as 
\begin{subequations}\label{llp_1}
	\begin{align}
		\mathop {\min }\limits_{{\bf{p}},{\bf{f}}, {\bf{v}}} \quad & E \label{} \\
		\st\ \quad  & \eqref{llp1}, \eqref{llp4}-\eqref{llp6}.
	\end{align}
\end{subequations}
However, this problem is still non-convex and intractable to derive optimal solutions due to the variables are closely coupled in the objective function and constraint \eqref{llp1}.
To effectively solve problem \eqref{llp_1}, the decomposition method is adopted by decomposing it into three subproblems and solving them iteratively.

\subsubsection{Optimization of DT mapping data ratio}
Since $v_n$ of each client is independent in problem \eqref{llp_1}, we can equivalently transform it into minimizing the energy consumption for each client $n$.
Given $p_n$ and $f_n$, the subproblem of optimizing $v_n$ can be formulated as follows:
\begin{subequations}\label{llp_v}
	\begin{align}
		\mathop {\min }\limits_{{{v_n}}} \quad & \frac{\tau }{2}{c_n}\left( {1 - {v_n}} \right){D_n}f_n^2 \label{llp_v_obj} \\
		\st\ \quad  & \frac{{{c_n}\left( {1 - {v_n}} \right){D_n}}}{{{f_n}}} \le A_n,  \label{llp_v_1} \\
		& 0 \le v_n \le {v_n^{\text{max}}}, \label{llp_v_2}
	\end{align}
\end{subequations}
where $A_n = T^{\text{max}} - t_n^{\text{com}}$ is regarded as a constant.
It can be observed that the objective function \eqref{llp_v_obj} is a monotonously decreasing function of $v_n$.
Thus, the optimal solution of problem \eqref{llp_v} depends on its constraints.
Combining constraints \eqref{llp_v_1} and \eqref{llp_v_2}, we can obtain the value limit of $v_n$ as
\begin{equation}\label{}
	\begin{aligned}
		1 - \frac{{{A_n}{f_n}}}{{{c_n}{D_n}}} \le {v_n} \le {v_n^{\text{max}}},
	\end{aligned}
\end{equation}
when $1 - \frac{{{A_n}{f_n}}}{{{c_n}{D_n}}} \le {v_n^{\text{max}}}$ is satisfied, which indicates that
\begin{equation}\label{con_fn}
	\begin{aligned}
		{f_n} \ge \frac{{\left( {1 - {v_n^{\text{max}}}} \right){c_n}{D_n}}}{{{A_n}}}.
	\end{aligned}
\end{equation}
In this case, combining the monotonicity of \eqref{llp_v_obj} and the value limit of $v_n$, the optimal solution is derived as $v_n^* = {v_n^{\text{max}}}$.
This indicates that the selected clients tend to map maximum portion of training data at the server via the DT network to minimize the energy consumption for local model training.

\subsubsection{Optimization of local computing frequency}
Similarly, due to the independence of $f_n$ in problem \eqref{llp_1}, the optimal solution can be obtained by minimizing the energy consumption of each client.
Given $p_n$ and $v_n$, we can reformulate the subproblem of optimizing $f_n$ as 
\begin{subequations}\label{llp_f}
	\begin{align}
		\mathop {\min }\limits_{{{f_n}}} \quad & \frac{\tau }{2}{c_n}\left( {1 - {v_n}} \right){D_n}f_n^2 \label{llp_f_obj} \\
		\st\ \quad  & \frac{{{c_n}\left( {1 - {v_n}} \right){D_n}}}{{{f_n}}} \le A_n,  \label{llp_f_1} \\
		& {f_n^{\text{min}}} \le f_n \le {f_n^{\text{max}}}. \label{llp_f_2}
	\end{align}
\end{subequations}
Note that from \eqref{llp_f_1}, we derive that ${f_n} \ge \frac{{\left( {1 - {v_n}} \right){c_n}{D_n}}}{{{A_n}}}$ which always satisfies \eqref{con_fn}.
It is easy to prove that the objective function \eqref{llp_f_obj} increases monotonously with respect to $f_n$ within the feasible range.
Thus, define ${\tilde f_n} = \frac{{\left( {1 - {v_n}} \right){c_n}{D_n}}}{{{A_n}}}$, and we can obtain the optimal solution of $f_n$ by analyzing following two cases:

If ${\tilde f_n} \le {f_n^{\text{min}}}$, the optimal solution is derived by $f_n^* = {f_n^{\text{min}}}$; 
Otherwise, if ${f_n^{\text{min}}} < {\tilde f_n} \le {f_n^{\text{max}}}$, the optimal solution is derived by $f_n^* = {\tilde f_n}$.
The analysis is illustrated in Fig. \ref{Solution_f}.
Combining these two cases, we can express the optimal solution in a uniform form as $f_n^* = \max \left\{ {{\tilde f_n}, {f_n^{\text{min}}}} \right\}$.

\begin{figure}[t]
	\centering
	\subfigure[Case 1: ${\tilde f_n} \le {f_n^{\text{min}}}$] { \label{Solution_f1}
		\includegraphics[width=0.47\columnwidth]{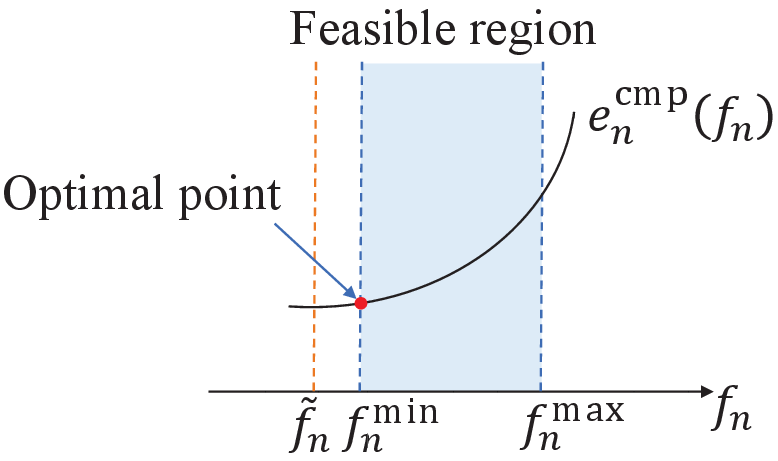}
	}
	\subfigure[Case 2: ${f_n^{\text{min}}} < {\tilde f_n} \le {f_n^{\text{max}}}$] { \label{Solution_f2}
		\includegraphics[width=0.45\columnwidth]{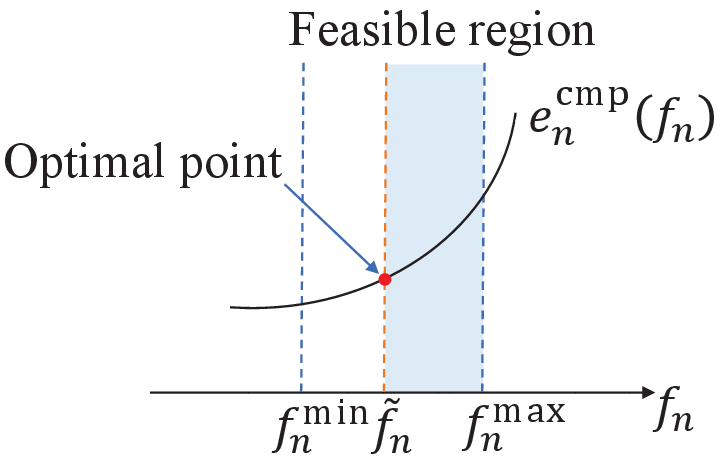}
	}
	\caption{Solution analysis of $f_n$. }
	\label{Solution_f}
\end{figure}

\subsubsection{Optimization of transmitting power}
Due to the simultaneous transmission of $N$ clients with NOMA, we cannot directly transform problem \eqref{llp_1} into $N$ subproblems to optimize the transmitting power by minimizing the energy consumption of each client.
However, based on the following analysis, the successive resource allocation mechanism \cite{SuccessivePower} can be utilized to address the problem.
Specifically, the use of SIC indicates that client $n$'s choice for transmitting power $p_n$ has no influence on the data rate of client $m$, where $n < m$.
Take the last decoded client, i.e., client $N$, as an extreme example, its data rate $R_N$ is only determined by its own transmitting power $p_N$ and is not influenced by the transmitting power of clients decoded before it.
Thus, $p_n$ can be optimized after $p_{n+1}$, a process known as successive optimization.
In this case, we can minimize the energy consumption of client $n$ by optimizing only $p_n$, because the transmitting power of client $m$, $n+1 \le m \le N$, has already been optimized, while the transmitting power of client $i$, $i \le n-1$, has no influence on its data rate.

Based on the above analysis, by fixing $v_n$ and $f_n$, the original problem \eqref{llp_1} can be transformed into the following $N$ subproblems, which are then be solved in a successive manner.
\begin{subequations}\label{llp_p}
	\begin{align}
		\mathop {\min }\limits_{{{p_n}}} \quad & \frac{{{p_n}{d_n}}}{{B{{\log }_2}\left( {1 + {p_n}{F_n}} \right)}} \label{llp_p_obj} \\
		\st\ \quad  & {B{{\log }_2}\left( {1 + {p_n}{F_n}} \right)} \le \frac{{d_n}}{G_n},  \label{llp_p_1} \\
		& {p_n^{\text{min}}} \le p_n \le {p_n^{\text{max}}}, \label{llp_p_2}
	\end{align}
\end{subequations}
where $F_n = \frac{{{{\left| {{h_n}} \right|}^2}}}{{\sum\limits_{j = n + 1}^N {{p_j}{{\left| {{h_j}} \right|}^2}}  + \sigma _n^2}}$ and $G_n = T^{\text{max}} - \frac{{{c_n}\left( {1 - {v_n}} \right){D_n}}}{{{f_n}}}$ are treated as constants.
However, the above problem is still non-convex due to the non-convexity of the fractional objective function \eqref{llp_p_obj} \cite{convex}.
To address it effectively, we first transform the minimization problem into the following equivalent maximization problem:
\begin{subequations}\label{llp_p_max}
	\begin{align}
		\mathop {\max }\limits_{{{p_n}}} \quad & \frac{{{B{{\log }_2}\left( {1 + {p_n}{F_n}} \right)}}}{{p_n}{d_n}} \label{llp_p_maxobj} \\
		\st\ \quad  & \eqref{llp_p_1}, \eqref{llp_p_2}.
	\end{align}
\end{subequations}
It is observed that the numerator and denominator of \eqref{llp_p_maxobj} are concave and convex functions with respect to $p_n$, respectively, which indicates that problem \eqref{llp_p_max} is a concave-convex fractional optimization problem.
Referring to \cite{FractionalForm}, by introducing an auxiliary parameter $q$, it can be equivalently transformed into the following more tractable form
\begin{subequations}\label{llp_p_max1}
	\begin{align}
		\mathop {\max }\limits_{{{p_n}}} \quad & Q\left( {p_n} \right) \buildrel \Delta \over = R\left( {p_n} \right) - qU\left( {p_n} \right) \label{llp_p_maxobj1} \\
		\st\ \quad  & \eqref{llp_p_1}, \eqref{llp_p_2},
	\end{align}
\end{subequations}
where $R\left( {p_n} \right) = {{B{{\log }_2}\left( {1 + {p_n}{F_n}} \right)}}$ and $U\left( {p_n} \right) = {{p_n}{d_n}}$.

We denote the optimal solution of problem \eqref{llp_p_max1} by ${\hat p_n}$ for a given $q$.
Define $p_n^*$ and $q^*$ as the optimal solution and maximum value of problem \eqref{llp_p_max}, respectively, i.e., ${q^ * } = R\left( {p_n^ * } \right)/U\left( {p_n^ * } \right) = \max \left\{ {R\left( {{p_n}} \right)/U\left( {{p_n}} \right)} \right\}$.
We formulate the following function as 
\begin{equation}\label{}
	\begin{aligned}
		W\left( q \right) \buildrel \Delta \over = \mathop {\max }\limits_{{p_n}} \left\{ {R\left( {{p_n}} \right) - qU\left( {{p_n}} \right)} \right\}.
	\end{aligned}
\end{equation}
As revealed in \cite{dinkelbach}, solving problem \eqref{llp_p_max} is equivalent to find a ${q^*}$ which satisfies 
\begin{equation}\label{}
	\begin{aligned}
			W\left( {{q^ * }} \right) & = \mathop {\max }\limits_{{p_n}} \left\{ {R\left( {{p_n}} \right) - {q^ * }U\left( {{p_n}} \right)} \right\}\\
			 & = R\left( {p_n^ * } \right) - {q^ * }U\left( {p_n^ * } \right) = 0.
	\end{aligned}
\end{equation}
The Dinkelbach algorithm can be designed to find the optimal ${q^*}$ in an iterative manner.
Specifically, at each iteration, we first obtain $\hat p_n^ {(j)}$ by solving problem \eqref{llp_p_max1} with a given ${q^{(j)}}$.
Next, ${q^{(j)}}$ is updated according to ${q^{(j+1)}} = {R\left( {\hat p_n^{\left( j \right)}} \right)} / {U\left( {\hat p_n^{\left( j \right)}} \right)}$.
As proved in \cite{dinkelbach}, if the problem \eqref{llp_p_max1} can be solved at each iteration, the above iterative process can always converge to the optimal ${q^*}$.
Note that once ${q^*}$ is found, the optimal solution ${p_n^ * }$ is also obtained.
The followings analyze the process of solving problem \eqref{llp_p_max1} at each iteration with given ${q}$.

It can easily be proved that \eqref{llp_p_max1} is a convex problem, which satisfies the Slater’s condition.
Thus, the Karush-Kuhn-Tucker (KKT) conditions can be exploited to derive the optimal solution.
The Lagrangian function of problem \eqref{llp_p_max1} can be written as
\begin{equation}\label{}
	\begin{aligned}
		{{\cal{L}}(p_n, {\bm{\lambda }})} & = {{B{{\log }_2}\left( {1 + {p_n}{F_n}} \right)}} -q{{p_n}{d_n}} \\
		& + {\lambda_1}\left(\frac{{d_n}}{G_n} -  {B{{\log }_2}\left( {1 + {p_n}{F_n}} \right)}\right) \\
		& + {\lambda_2}\left( {p_n^{\text{min}}} - p_n\right) + {\lambda_3}\left( p_n - {p_n^{\text{max}}} \right),
	\end{aligned}
\end{equation}
where ${\lambda_i}$ are Lagrangian multipliers related to the constraints in problem \eqref{llp_p_max1}.
The dual function can be expressed as 
\begin{equation}\label{}
	\begin{aligned}
		g \left( {\bm{\lambda }} \right) = \mathop {\max }\limits_{{p_n}} {\cal{L}}\left( {{p_n},{\bm{\lambda }}} \right).
	\end{aligned}
\end{equation}
Taking the first-order partial derivative of ${\cal{L}}(p_n, {\bm{\lambda }})$ with $p_n$, we have 
\begin{equation}\label{}
	\begin{aligned}
		\frac{{\partial {\cal{L}}}}{{\partial {p_n}}} &= \frac{{B{F_n}}}{{\ln 2\left( {1 + {p_n}{F_n}} \right)}} - q{d_n} \\
		&- \frac{{{\lambda _1}B{F_n}}}{{\ln 2\left( {1 + {p_n}{F_n}} \right)}} - {\lambda _2} + {\lambda _3}.
	\end{aligned}
\end{equation}
Let $\frac{{\partial {\cal{L}}}}{{\partial {p_n}}} = 0$ and we can obtain the optimal solution as 
\begin{equation}\label{Opt_power}
	\begin{aligned}
		p_n^ *  = \frac{{B\left( {1 - {\lambda _1}} \right)}}{{\ln 2\left( {q{d_n} + {\lambda_2} - {\lambda _3}} \right)}} - \frac{1}{{{F_n}}}.
	\end{aligned}
\end{equation}

To obtain the Lagrangian multipliers ${\bm{\lambda }}$, we first formulate the dual problem as 
\begin{subequations}\label{}
	\begin{align}
		\mathop {\min }\limits_{{{\bm{\lambda }}}} \quad & g \left( {\bm{\lambda }} \right) \label{} \\
		\st\ \quad  & {\bm{\lambda }} \succeq  0.
	\end{align}
\end{subequations}
Adopting the subgradient method \cite{TSN}, the dual problem can be effectively addressed.
The Lagrangian multipliers are updated according to
\begin{subequations}\label{}
	\begin{align}
		&\lambda _1^{\left( {l + 1} \right)} = {\left( {\lambda _1^{\left( l \right)} - {\mu _1} \left( {\frac{{{d_n}}}{{{G_n}}} - B{{\log }_2}\left( {1 + {p_n}{F_n}} \right)} \right)} \right)^ + },\\
		&\lambda _2^{\left( {l + 1} \right)} = {\left( {\lambda _2^{\left( l \right)} - {\mu _2}\left( {{p_n^{\text{min}}} - p_n} \right)} \right)^ + },\\
		&\lambda _3^{\left( {l + 1} \right)} = {\left( {\lambda _3^{\left( l \right)} - {\mu _3}\left( {p_n - {p_n^{\text{max}}}} \right)} \right)^ + },
	\end{align}
\end{subequations}
where $l$ and ${\mu _i}$ denote the iteration index and updating step size, respectively.
Algorithm \ref{algorithm1} summarizes the process of transmitting power optimization based on the Dinkelbach approach.

\begin{algorithm}[t]
	\footnotesize
	\caption{Optimization of Transmitting Power.}
	\label{algorithm1}
	\begin{algorithmic}[1] 
		\STATE Initialize ${q^{(0)}} = 0$, ${W^{(0)}} = \infty $; set precision $\delta > 0$.
		\STATE \textbf{While} $\left| {W\left( q \right)} \right| > \delta $ \textbf{do}
		\STATE \qquad Obtain optimal $p_n^ *$ according to \eqref{Opt_power};
		\STATE \qquad Calculate ${W\left( q \right)}$;
		\STATE \qquad Update ${q} = {R\left( p_n^*\right)} / {U\left( p_n^* \right)}$;
		\STATE \textbf{Return} $q$ and $p_n^ * $.
	\end{algorithmic}
\end{algorithm}

\subsection{Overall Algorithm Design}
\begin{algorithm}[t]
	\footnotesize
	\caption{Joint Optimization of Follower-Level and Leader-Level Problems.}
	\label{algorithm2}
	\begin{algorithmic}[1] 
		\STATE Initialize variables $\alpha_n$, $p_n$, $f_n$ and $v_n$ for each client $n \in {\cal N}$.
		\STATE \textbf{Repeat} 
		\STATE  \qquad \textbf{The follower:} 
		\STATE  \qquad Obtain strategies of $\alpha_n^*$ according to \eqref{Solution_a} and \eqref{OptAl}.
		\STATE  \qquad \textbf{The leader $n \in \left\{ N, N-1,..., 1\right\} $:} 
		\STATE  \qquad \textbf{Repeat} 
		\STATE  \qquad \qquad Set $v_n^* = {v_n^{\text{max}}}$;
		\STATE  \qquad \qquad Calculate $f_n^* = \max \left\{ {{\tilde f_n}, {f_n^{\text{min}}}} \right\}$;
		\STATE  \qquad \qquad Obtain $p_n^*$ from Algorithm \ref{algorithm1}.
		\STATE  \qquad \textbf{Until $E$ converges} 
		\STATE \textbf{Until realize Stackelberg equilibrium}
		\STATE \textbf{Output:} $\alpha_n^*$, $v_n^*$, $f_n^*$ and $p_n^*$.
	\end{algorithmic}
\end{algorithm}

The overall algorithm to jointly solve the follower-level and leader-level problems is presented in Algorithm \ref{algorithm2}, which begins with the initialization of optimization variables.
In the main loop, the follower, i.e., the server, first obtains its optimal strategies of $\alpha_n^*$ according to \eqref{Solution_a} and \eqref{OptAl}. 
Then, for the leader-level problem, following the successive optimization mechanism, each client $n \in \left\{ N, N-1,..., 1\right\} $ can optimize its resource allocation variables in closed-form expressions iteratively.
This optimization loop proceeds until the Stackelberg equilibrium is realized.
Thus, the optimal solutions to minimize the total latency and energy consumption can be obtained effectively.

\begin{table}[t]
	\footnotesize
	\begin{center}
		\caption{\protect\\\textsc{Simulation Parameters}}\vspace{+1em}
		\label{Para}
		\begin{tabular}{c|c}
			\hline
			Parameter & Value\\  \hline
			Carrier frequency & $1$ GHz\\
			Bandwidth, $B $ & $1$ MHz \\
			Path loss exponent & $3.76$\\
			AWGN spectral density & $-174$ dBm/Hz\\
			Transmit power limit, $p_n^{\text{min}}$, $p_n^{\text{max}}$ & $[0.01, 0.1] $  W\\
			CPU cycles for each sample, $c_n$ & $10^7$ \\
			Computation frequency limit, $f_n^{\text{min}}$, $f_n^{\text{max}}$ & $[1, 10]$ GHz\\
			Computation frequency of server, $f_S$ & $100$ GHz \\
			Maximum latency limit, ${T^{\max}}$ & $10$ s \\
			Local model size, $d_n $ & $ 1$ Mbit\\
			Learning rate, $\eta$ & $0.01$ \\
			\hline
		\end{tabular}
	\end{center}
\end{table}

\section{Simulation and Analysis}
In this section, extensive numerical simulations are conducted to evaluate the performance of the proposed scheme.
We consider a circular area of $500$ radius, where the server is deployed at the center and $20$ clients are distributed randomly.
Due to the inherent communication limitations, only $5$ clients are selected at each global FL round.
Considering the computing limitations at clients and the dynamic communication environment, the DT network is deployed at the server to assist FL training. 
Besides, we also consider several poisoners during FL process which cause label-flipping attacks to degrade FL performance \cite{PI}.
The FL models are trained using both the MNIST and CIFAR-10 datasets, considering both independent and identically distributed (IID) and non-IID data distributions for each dataset.
Specifically, for the IID data distribution, the training data and corresponding labels are identically distributed among all clients, though the data sizes may vary.
For the non-IID data distribution, each client has $1$ type of label in the MNIST dataset and $5$ types of labels in the CIFAR-10 dataset, respectively.
The other simulation parameters are presented in Table \ref{Para}.

\begin{figure}[t]
	\centering
	\includegraphics[width=0.45\textwidth]{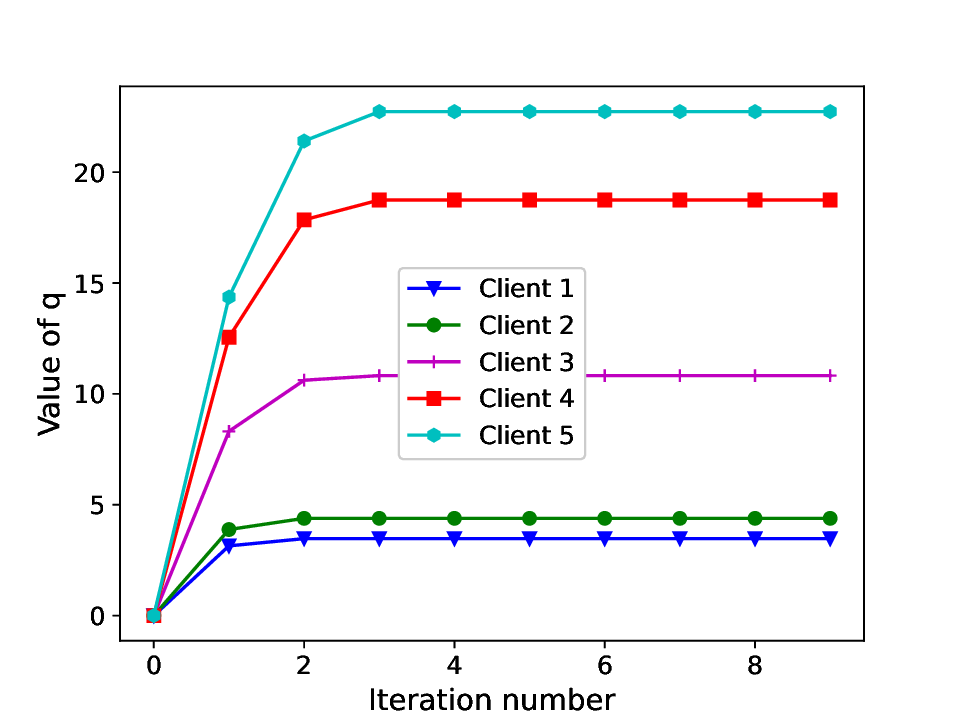}
	\caption{Convergence of Algorithm \ref{algorithm1}.}
	\label{Qvalue_iter}
\end{figure}


Fig. \ref{Qvalue_iter} illustrates the convergence performance of the proposed Algorithm \ref{algorithm1}, which adopts the Dinkelbach approach to optimize the transmitting power of selected clients in a successive manner.
The initial $q$ value is set as $0$, and it increases with the growth of iteration number for each client.
It can be found that Algorithm \ref{algorithm1} can converge to the optimal value in several iterations, which guarantees that the optimal transmitting power of selected clients can always be obtained.
Besides, following the successive optimization method, clients' $q$ values are related to their decoding order, with the first client to decode having the smallest $q$ value.


\subsection{FL performance with different number of poisoners} 

\begin{figure}[t]
	\centering
	\begin{minipage}[]{1\textwidth}
		\subfigure[FL accuracy on MNIST dataset]
		{\label{FL_0610_mni_iid}
			\includegraphics[width=0.45\textwidth]{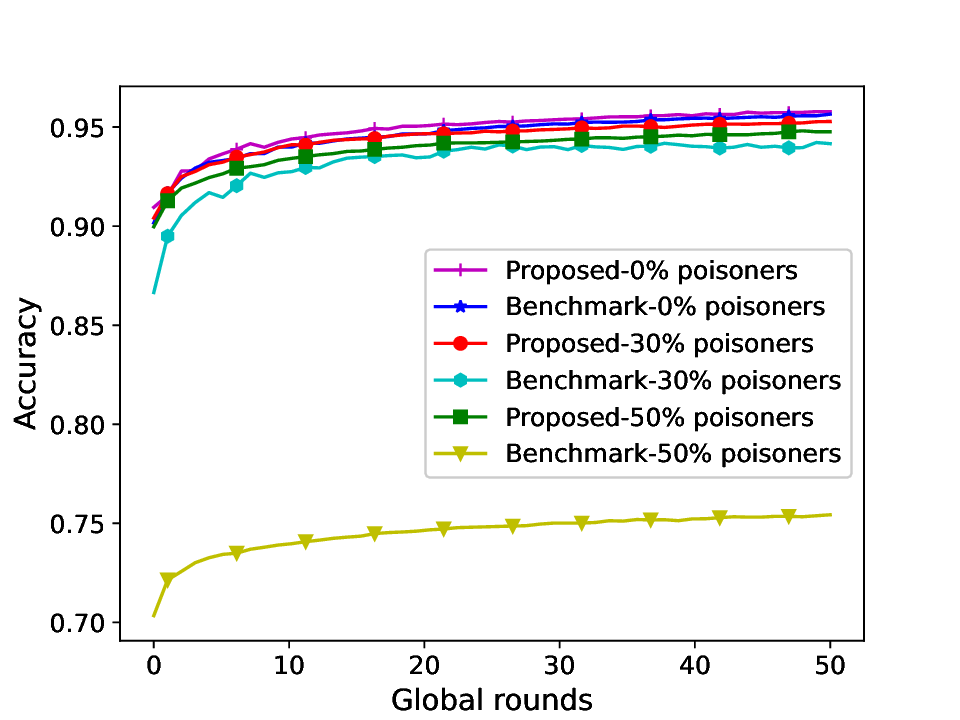}}
	\end{minipage}
	\begin{minipage}[]{1\textwidth}
		\subfigure[FL accuracy on CIFAR-10 dataset]
		{\label{FL_0610_saf_iid}
			\includegraphics[width=0.45\textwidth]{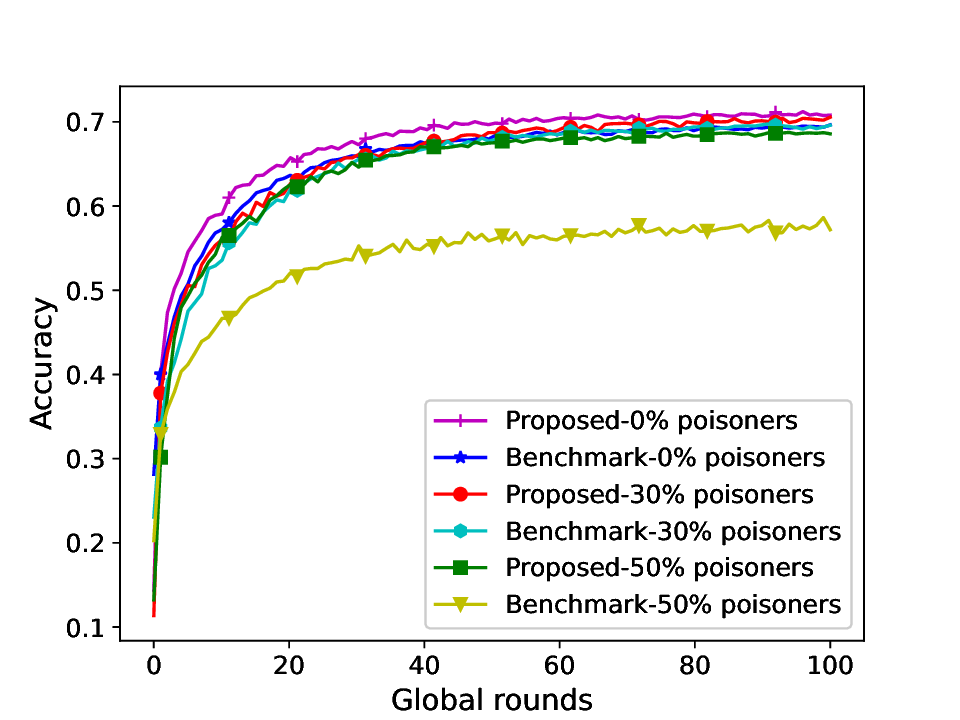}}
	\end{minipage}
	\caption{FL performance with different number of poisoners.}
	\label{FL_poisoners}
\end{figure}

In Fig. \ref{FL_poisoners}, the FL performance with different number of poisoners in both IID MNIST and CIFAR-10 datasets are illustrated.
The benchmark algorithm is introduced for comparison with the proposed scheme. 
In the benchmark algorithm, client reputation for selection considers only the factors of AC and MS, each with a weight of $0.5$, while ignoring the degree of PI. 
Consequently, the benchmark scheme is more vulnerable to poisoners.
In the proposed scheme, the weights for AC, MS and PI used to calculate the reputation value of clients are set to $0.3$, $0.5$ and $0.2$, respectively.
As shown in Fig. \ref{FL_poisoners}\subref{FL_0610_mni_iid}, when training on the MNIST dataset, the proposed and benchmark schemes with $0\%$ poisoners achieve nearly identical FL accuracy performance.
When the ratio of poisoners in the system is $30\%$, the proposed scheme outperforms the benchmark algorithm in FL accuracy performance.
This is because the proposed scheme evaluates the quality of local model updates before global model aggregation, thereby avoiding malicious attacks from poisoners.
Additionally, with $50\%$ poisoners, the proposed scheme shows a significant improvement in FL performance compared to the benchmark, which is severely impacted by the high portion of poisoners.

Fig. \ref{FL_poisoners}\subref{FL_0610_saf_iid} illustrates the FL accuracy on CIFAR-10 dataset.
It is observed that the proposed scheme with $0\%$ poisoners realizes the highest FL accuracy, followed by the proposed scheme with $30\%$ poisoners.
It demonstrates the effectiveness of the proposed scheme in enhancing FL performance under poisoning scenarios.
The benchmark scheme with $0\%$ shows inferior performance compared to its counterpart, which is attributed to the weights assigned to the different factors in the clients' reputation.
Obviously, the benchmark with $50\%$ poisoners exhibits the worst FL performance due to the severe poisoning attacks on the global model.

\subsection{FL performance with different DT deviations} 

\begin{figure}[t]
	\centering
	\begin{minipage}[]{1\textwidth}
		\subfigure[FL accuracy on MNIST dataset]
		{\label{FL_error_mni}
			\includegraphics[width=0.45\textwidth]{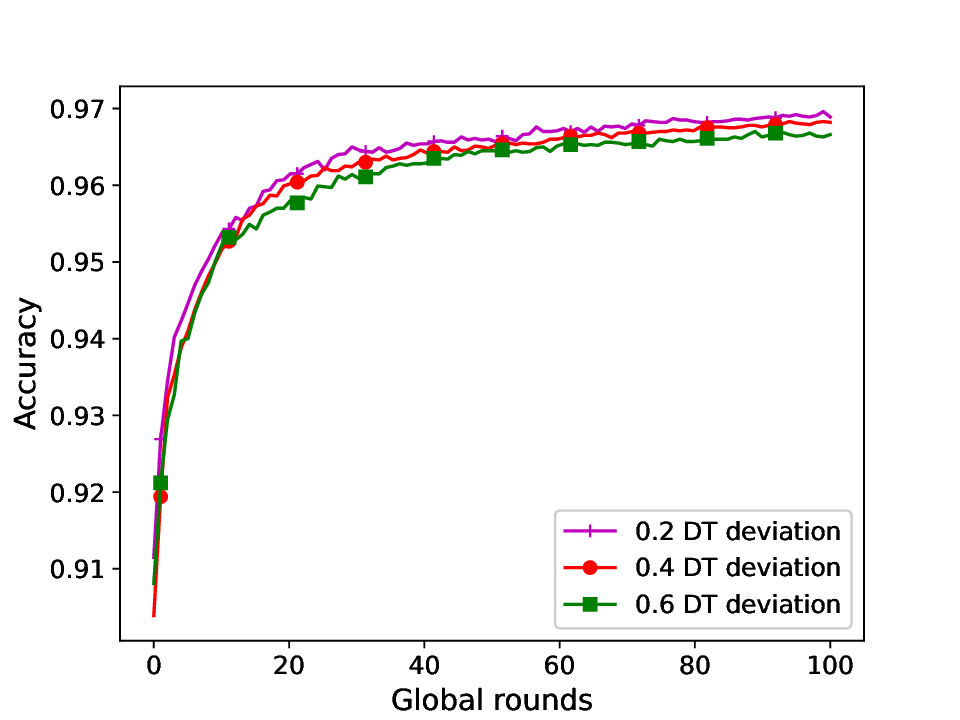}}
	\end{minipage}
	\begin{minipage}[]{1\textwidth}
		\subfigure[FL accuracy on CIFAR-10 dataset]
		{\label{FL_error_caf}
			\includegraphics[width=0.45\textwidth]{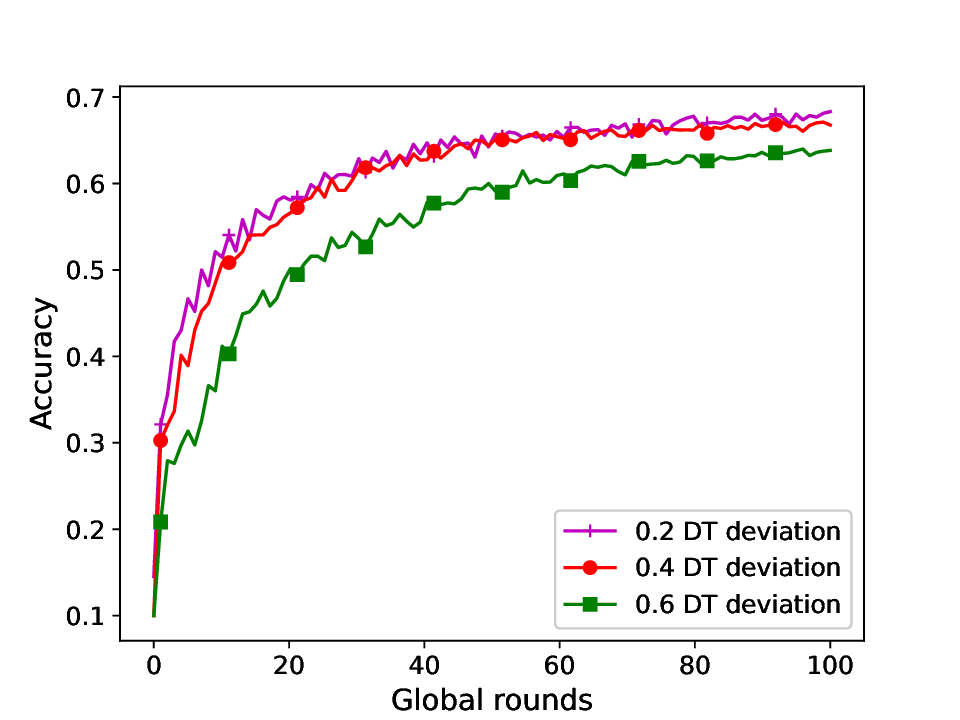}}
	\end{minipage}
	\caption{FL performance with different DT deviations.}
	\label{FL_error}
\end{figure}

Fig. \ref{FL_error}\subref{FL_error_mni} and Fig. \ref{FL_error}\subref{FL_error_caf} present the FL performance with different DT deviations for mapping data on IID MNIST and CIFAR-10 datasets, respectively.
Note that the DT deviation needs to be multiplied by a random value between $-1$ and $1$ before applying it to each mapping data.
Intuitively, as the DT deviation for mapping data increases, the FL accuracy decreases for both datasets.
This is because the larger DT deviation enlarges the difference between the real training data and the estimated one, thereby degrading the FL performance.
Besides, it can be found that for the MNIST dataset, the gap in FL accuracy among schemes with different DT deviations is small, which demonstrates its robustness to DT mapping error.
On the contrary, when the DT deviation is large, i.e., $0.6$, the FL performance on CIFAR-10 dataset decrease significantly due to its greater complexity compared to the MNIST dataset.
Thus, the more complex training dataset is more sensitive to the DT deviation for mapping data.

\begin{figure}[t]
	\centering
	\begin{minipage}[]{1\textwidth}
		\subfigure[FL accuracy on IID dataset]
		{\label{FL_mni_iid}
			\includegraphics[width=0.45\textwidth]{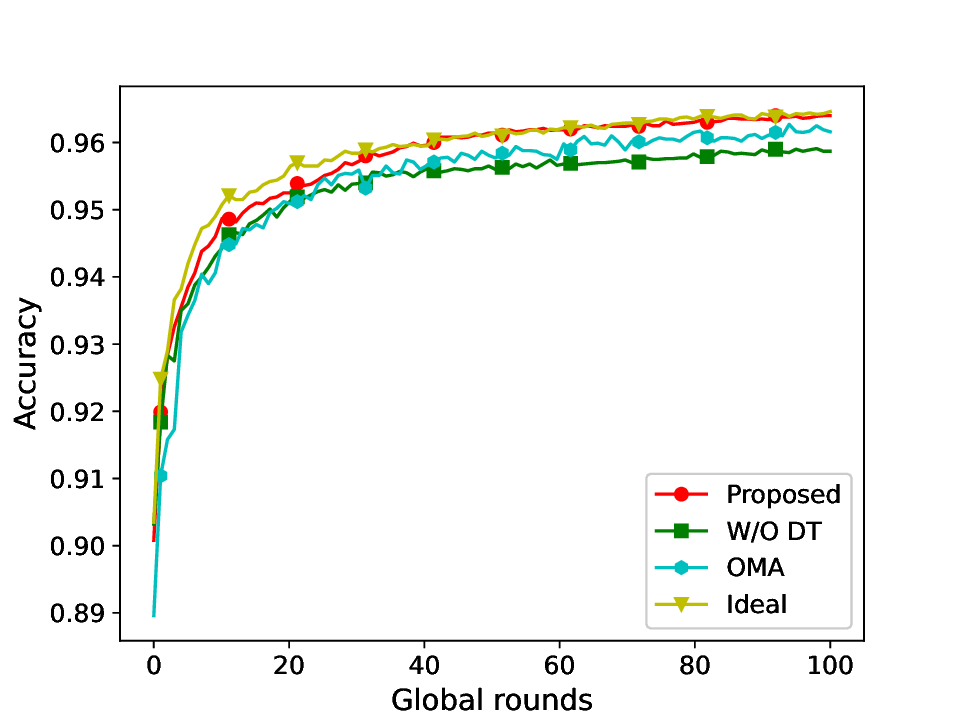}}
	\end{minipage}
	\begin{minipage}[]{1\textwidth}
		\subfigure[FL accuracy on non-IID dataset]
		{\label{FL_mni_niid}
			\includegraphics[width=0.45\textwidth]{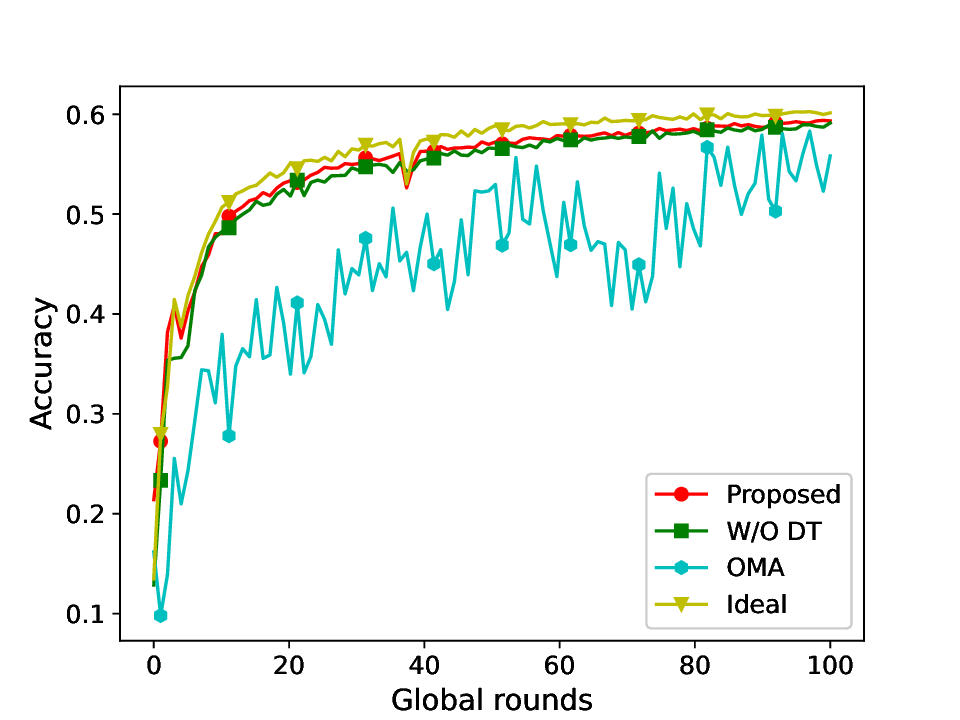}}
	\end{minipage}
	\caption{FL performance on MNIST dataset ($30\%$ poisoners).}
	\label{FL_MNIST}
\end{figure}

\begin{figure}[t]
	\centering
	\begin{minipage}[]{1\textwidth}
		\subfigure[FL accuracy on IID dataset]
		{\label{FL_saf_iid}
			\includegraphics[width=0.45\textwidth]{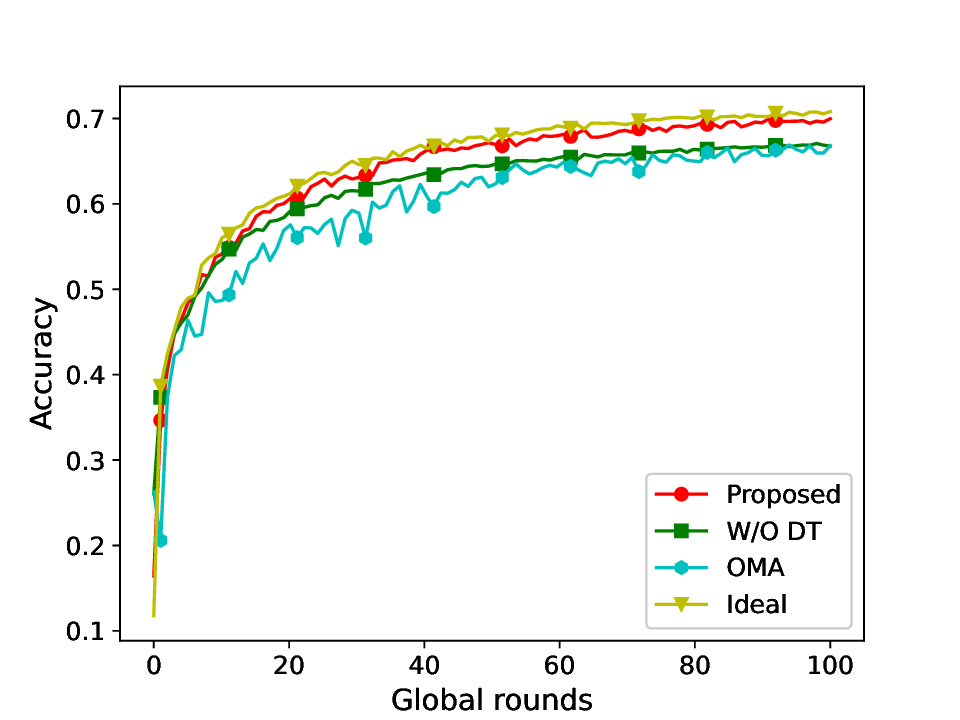}}
	\end{minipage}
	\begin{minipage}[]{1\textwidth}
		\subfigure[FL accuracy on non-IID dataset]
		{\label{FL_saf_niid}
			\includegraphics[width=0.45\textwidth]{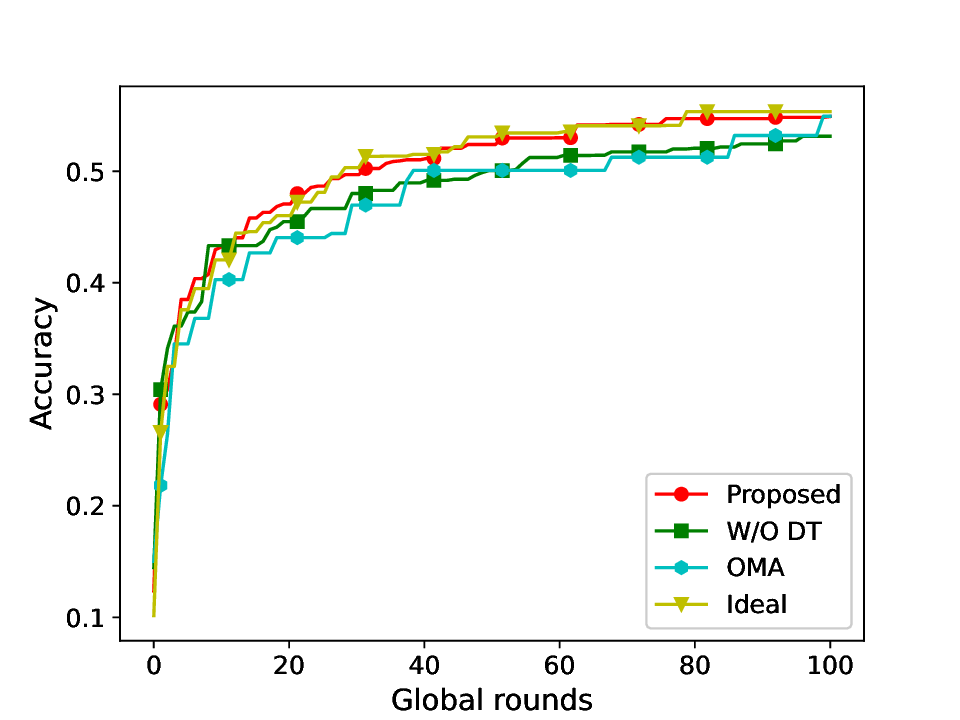}}
	\end{minipage}
	\caption{FL performance on CIFAR-10 dataset ($30\%$ poisoners).}
	\label{FL_CIFAR}
\end{figure}

\subsection{FL performance among different schemes}
In this subsection, to evaluate the performance of the proposed DT-assisted FL over NOMA scheme, several benchmarks are considered for comparison.
Specifically, in the FL without DT (W/O DT) scheme, the DT network is not deployed at the server, thereby clients suffer from issues of limited local computation resources and the dynamic communication environment. 
For the OMA scheme, the DT-assisted FL is conducted over OMA networks.
In the ideal case, we assume that clients have infinite computation resources to perform local model training so that the deployment of DT is not necessary.
Note that the reputation-based client selection is adopted in all the above schemes to ensure fairness in comparison.
We consider $30\%$ poisoners in the system, and both the MNIST and CIFAR-10 datasets are utilized for verification.

Fig. \ref{FL_MNIST}\subref{FL_mni_iid} and Fig. \ref{FL_MNIST}\subref{FL_mni_niid} present the FL performance on IID and non-IID MNIST dataset, respectively.
In the IID case, all schemes can achieve over $95\%$ FL accuracy.
We also find that the performance of the proposed scheme outperforms the W/O DT and OMA schemes and nearly reaches the performance of the ideal scheme.
It demonstrates that the deployment of DT at the server is beneficial for enhancing FL performance under practical computation and communication limitations.
Nevertheless, in the non-IID case, the FL accuracy drops significantly due to the unbalanced data distribution and the presence of poisoners in the system.
The proposed scheme still shows superior performance compared to the W/O DT and OMA schemes.
In particular, the OMA scheme has the worst performance and is not robust, due to the insufficient selected clients at each round.
Hence, NOMA is promising in improving FL performance given the inherent communication constraints.

Fig. \ref{FL_CIFAR}\subref{FL_saf_iid} illustrates the FL performance on IID CIFAR-10 dataset.
It is observed that the proposed scheme always realizes better FL accuracy performance compared with the W/O DT and OMA schemes.
This is because the deployment of DT can assist FL training when the local model training and uploading fail due to insufficient local computation resources and dynamic communication environment.
Besides, NOMA transmission achieves enhanced spectrum efficiency than OMA scheme.
Note that there exists an inherent small gap between the proposed scheme and the ideal case since the DT mapping introduces deviations between the real data and the estimated one.
The FL performance on non-IID CIFAR-10 dataset is shown in Fig. \ref{FL_CIFAR}\subref{FL_saf_niid}.
For this non-IID case, the FL performance is displayed using the maximum accuracy to avoid overcrowding among different algorithms \cite{KaidiFL}.
From Fig. \ref{FL_CIFAR}\subref{FL_saf_niid}, we observe that the proposed scheme achieves superior FL performance than the W/O DT and OMA schemes, and can approach to the performance of ideal case.
This demonstrates the effectiveness of introducing DT in FL to address the straggler issue.

\subsection{Performance of total cost}

\begin{figure*}[t]
	\centering
	\subfigure[Total cost versus $d_n$] { \label{Cost_dn}
		\includegraphics[width=0.31\textwidth]{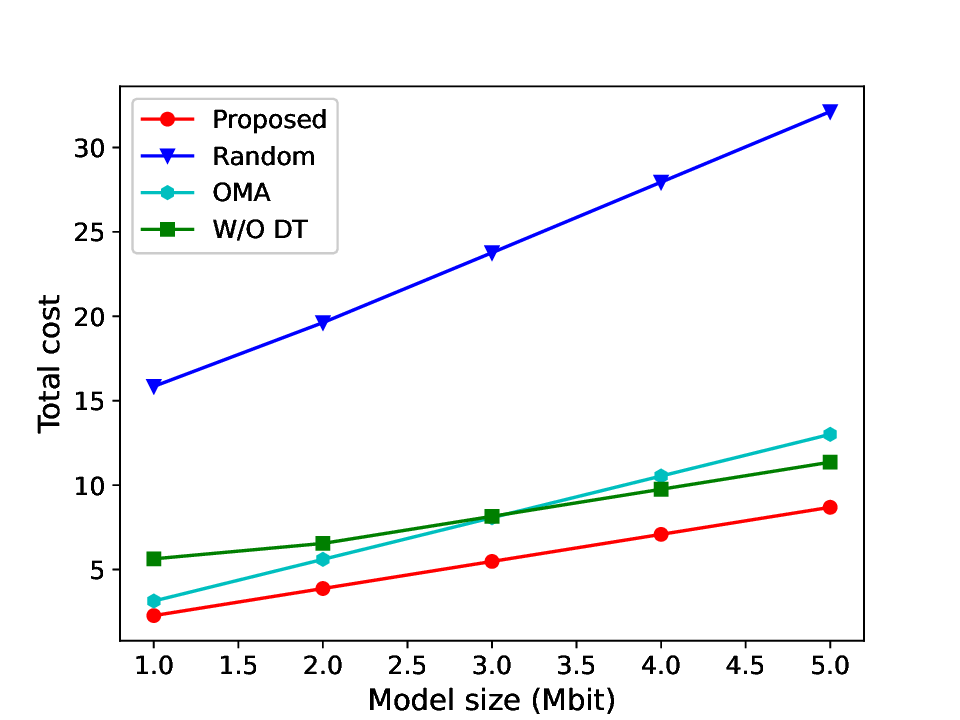}
	}
	\subfigure[Total cost versus $N$] { \label{Cost_Nm}
		\includegraphics[width=0.31\textwidth]{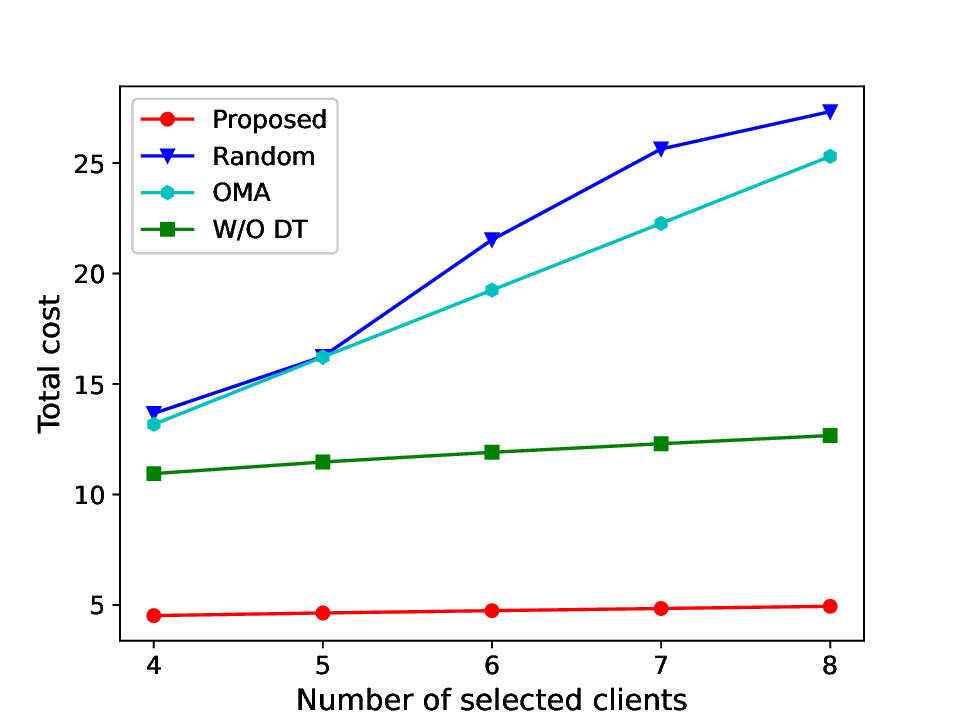}
	}
	\subfigure[Total cost versus $B$] { \label{Cost_B}
		\includegraphics[width=0.31\textwidth]{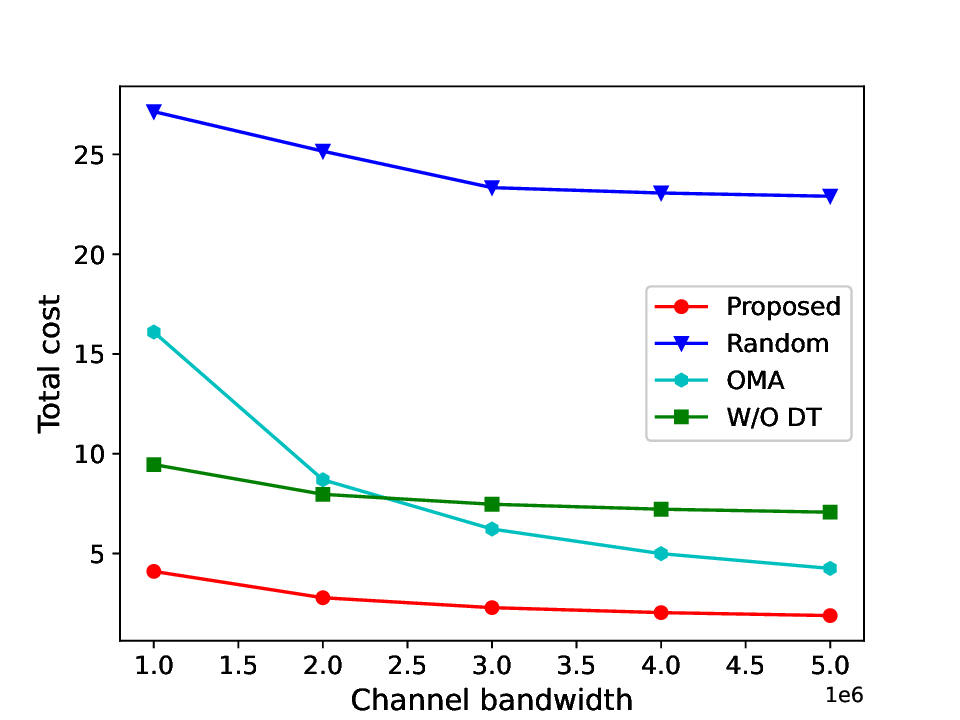}
	}
	\caption{Performance of total cost.}
	\label{TotalCost}
\end{figure*}

The performance of total cost, i.e., total latency and energy consumption, across different schemes is presented in Fig. \ref{TotalCost}.
The random scheme indicates that resources are allocated to the selected clients in a random manner, while other settings are the same with the proposed scheme.
As shown in Fig. \ref{TotalCost}\subref{Cost_dn}, the total cost increases with the growth of local model size $d_n$, as the larger $d_n$ contributes to increased wireless transmission latency for global model aggregation and corresponding energy consumption.
Besides, the proposed scheme realizes the lowest total cost compared with the other three benchmarks.
This can be explained from two aspects: first, the deployed DT network at the server assists the FL process, thereby reducing the cost associated with clients' local model training; second, NOMA achieves better communication efficiency compared to the OMA scheme, resulting in a reduced total cost.

In Fig. \ref{TotalCost}\subref{Cost_Nm}, as the number of selected clients $N$ increases, the total cost of the random and OMA schemes rises significantly owing to the randomness in resource allocation and low communication efficiency, respectively.
Since NOMA allows multiple clients to transmit on the same channel simultaneously, the total cost for the proposed scheme and the W/O DT scheme increases slowly as $N$ grows.
However, due to the complexity of SIC in NOMA, the appropriate number of selected clients $N$ should be carefully considered in practice.
Moreover, the proposed scheme still outperforms the others in terms of total cost, which validates its effectiveness in reducing total cost.

Fig. \ref{TotalCost}\subref{Cost_B} illustrates the performance of total cost versus available channel bandwidth $B$.
As $B$ increases, the total cost of all schemes decreases rapidly and then gradually stabilizes.
This is because the available transmitting rate is increased with larger $B$, thereby decreasing the wireless transmitting latency and corresponding energy consumption for uploading local model parameters.
However, further increasing $B$ does not significantly reduce the total cost and leads to a waste of communication resources.
It can also be observed that the proposed scheme consistently demonstrates superior performance compared to its counterparts.
Thus, the integration of DT and NOMA, along with the proposed resource allocation solutions, is effective to achieve low total cost during global model training for FL systems.

\section{Conclusion}
In this paper, we studied the minimization of latency and energy consumption in the DT-assisted FL system over NOMA network, while considering the presence of unreliable clients.
DT was deployed at the server to alleviate the straggler issue of FL.
A reputation-based client selection scheme was proposed to guarantee FL performance under malicious attacks from clients, which simultaneously accounts for multiple aspects of client heterogeneity, including clients' accuracy contribution, local model staleness, and positive interactions.
We considered the competitive interaction between clients and the server which are regarded as the leader and follower respectively, and formulated a Stackelberg game to achieve the minimization of their individual objectives. 
To obtain the optimal solutions, the game equilibrium was considered.
The solutions of the follower-level problem were first obtained through revealed insights,  which were then incorporated into the leader-level problem, allowing us to derive closed-form solutions via problem decomposition.
The simulation results demonstrated that our proposed scheme has superior performance in improving FL performance and reducing total cost, even in the unreliable training environment.

\bibliographystyle{IEEEtran}
\bibliography{EEref}
\vspace{0.5em}

\end{document}